\documentclass[11pt]{article}

\usepackage[final]{acl}

\usepackage{times}
\usepackage{latexsym}
\usepackage{graphicx}
\usepackage{xcolor,colortbl}
\usepackage{soul}
\usepackage[normalem]{ulem}
\usepackage{setspace}
\usepackage{balance}
\usepackage{float}
\usepackage{placeins}
\usepackage{url}

\usepackage{booktabs}
\usepackage{multirow}
\usepackage{makecell}
\usepackage{hhline}
\usepackage{threeparttable}
\usepackage{arydshln}
\usepackage{tabularray}
\usepackage{tabularx}
\usepackage{siunitx}

\usepackage[inline]{enumitem}

\usepackage{wrapfig}
\usepackage{subfig}
\usepackage{caption}

\usepackage{pgfkeys}
\usepackage{tikz}

\usepackage[most]{tcolorbox}
\tcbuselibrary{skins,breakable,listings}

\usepackage{pifont}
\newcommand{\cmark}{\ding{51}}
\newcommand{\xmark}{\ding{55}}
\newcommand{\pmark}{\(\triangle\)}

\usepackage{algorithm}
\usepackage{algorithmic}

\usepackage{lscape}

\usepackage{hyperref}
\usepackage{bookmark}
\usepackage{etoc}

\usepackage{graphicx}
\usepackage{amsmath}

\usepackage{amsfonts}

\usepackage[T1]{fontenc}

\usepackage[utf8]{inputenc}

\usepackage{microtype}

\usepackage{inconsolata}

\usepackage{graphicx}

\title{AgriGym: Benchmarking Evidence-Grounded Tool-Use in Multimodal Agents for Agricultural Consultations}

\author{Boyu Luo, Yilun Liu, Yanran Tang, Ruihong Qiu and Zi Huang \\
  School of Electrical Engineering and Computer Science \\
  The University of Queensland \\
  Brisbane, Queensland, Australia \\
  \texttt{boyu.luo@uq.edu.au}}

\begin{document}
\maketitle
\begin{abstract}
Agricultural AI assistants are increasingly expected to support plant health consultation. Current agricultural benchmarks mainly evaluate agricultural knowledge understanding or consultation response generation, but they largely overlook the evidence-seeking process used by agricultural experts.
Real agricultural consultation is a problem-solving process under uncertainty, where experts gather evidence, compare possible causes, and justify recommendations with supporting rationales.
To address this gap, we introduce \textbf{AgriGym}, an executable benchmark that models agricultural consultation as a multi-step agent-environment interaction.
In AgriGym, an agent repeatedly observes the current task state, requests additional evidence, evaluates the returned feedback, and eventually produces an evidence-grounded final recommendation.
Since agricultural consultation does not admit a single gold reasoning path, AgriGym does not evaluate agents by matching predefined tool sequences. 
Instead, it assesses whether an agent can identify missing evidence, retrieve relevant information, and ground its final recommendation in the evidence collected throughout the consultation process. The trajectory-aware judge separately assesses outcome quality and process reliability by examining evidence grounding, trajectory quality, visual evidence use, and tool-use efficiency. Human-expert verification and cross-judge evaluation provide preliminary evidence of alignment with expert assessment and alternative judges.
Experiments with diverse vision-language agents show that AgriGym enables comprehensive evaluation of agricultural agents by separating final recommendation quality from the reliability of the supporting evidence-seeking process. The code is available via \url{https://github.com/WilfridLuo/agri-gym}.
\end{abstract}

\section{Introduction}
Agricultural consultation is critical for crop health management and sustainable farming, yet access to timely expert support is often limited by geographic and scalability constraints. Recent vision-language models (VLMs)~\cite{llava,ds-vl2,Qwen3-VL,qwen2.5vl,gemini2.5, internvl3.5} have shown strong capabilities in jointly understanding agricultural images and natural-language consultation queries, making them promising foundations for agricultural AI assistants. Consequently, recent agricultural benchmarks have focused on evaluating multimodal agricultural understanding, crop diagnosis, and consultation response generation~\cite{agrobench,agricot-bench,agmmu,mirage}.

\begin{figure}[!t]
\centering
\includegraphics[width=1\linewidth]{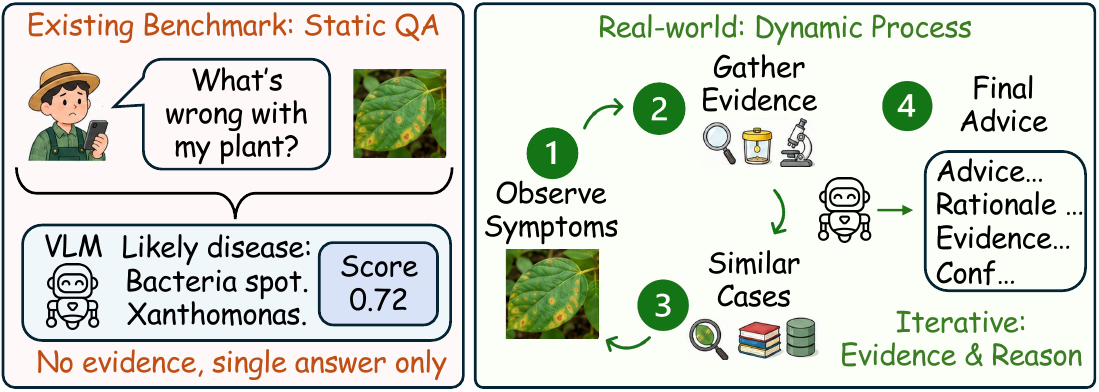}
\caption{\textbf{Agricultural assistant: Static QA versus dynamic consultation.}
Existing benchmarks usually evaluate agricultural assistant via \textbf{static QA} (Left), where only the final answer is scored. Real consultation is a \textbf{dynamic problem solving process} (Right). Experts gather supporting evidence from field photos, example cases, then compare the current symptoms with similar cases and knowledge bases. The final advice is therefore grounded in observed symptoms, external evidence, and accumulated reasoning, motivating evaluation beyond final answer only.
}
\label{fig:motivation}
\vspace{-0.5cm}
\end{figure}

Existing agricultural benchmarks mainly evaluate visual recognition, agricultural knowledge understanding~\cite{crop, cddm, agrogpt, agrigpt-rag}, and expert-style consultation response generation~\cite{ecosage,mirage,agrieval}. Most of these settings still formulate agricultural assistance as a static question-answering task, where a model receives an agricultural query with optional visual context and is evaluated primarily based on its final response. Such outcome-centric evaluation provides limited insight into whether the model reaches its recommendation through a credible evidence-seeking process that supports the final recommendation.

In contrast, real agricultural consultation is rarely a single-step response-generation problem. When assessing a plant health case, experts typically form an initial diagnosis from observed symptoms, gather additional evidence to support or eliminate possible causes, refine their assessment, and finally produce an evidence-grounded recommendation. This process is particularly important in agriculture because visually similar symptoms may arise from fundamentally different causes, including diseases, pest infestations, nutrient deficiencies, and environmental stress conditions. As illustrated in Figure~\ref{fig:motivation}, existing response-level benchmarks therefore remain poorly aligned with the evidence-seeking workflow underlying real agricultural consultation. This motivation is consistent with recent interactive agent frameworks, which evaluate models through environment interaction rather than isolated outputs~\cite{agentbench,webarena,react}. Moreover, such evaluation cannot rely on exact action matching, since multiple evidence-seeking trajectories may reasonably lead to similarly credible conclusions.

To address this limitation, we introduce AgriGym, an executable benchmark framework for evidence-grounded agricultural consultation agents. AgriGym reformulates agricultural consultation from static response generation into a multi-step evidence-seeking process.
Agents must gather information, compare candidate causes, and ground their final recommendations in accumulated evidence.
Within this process, tool use serves as the mechanism for acquiring and inspecting evidence, rather than as an auxiliary interaction detached from the final answer. AgriGym further introduces a trajectory-aware evaluation framework for assessing consultation reliability without requiring exact action matching. Because agricultural consultation does not admit a single gold reasoning path, the evaluation instead examines whether the collected evidence supports the agent's final recommendation.
The trajectory-aware judge separately assesses outcome quality and process reliability by examining evidence grounding, trajectory quality, visual evidence use, and tool-use efficiency.
Experiments across diverse vision-language agents demonstrate that AgriGym provides a more diagnostic evaluation of agricultural agents than response-level benchmarks by separating final recommendation quality from the reliability of the supporting evidence-seeking process. Our contributions are summarized as follows:
\begin{itemize}
\item We introduce \textsc{AgriGym}, an executable benchmark that reformulates static agricultural consultation into multi-step agent--environment interaction with explicit evidence gathering and tool use.

\item We propose a trajectory-aware evaluation protocol that separately assesses final recommendation quality and process reliability through evidence grounding, trajectory quality, tool-use efficiency, and visual evidence use.

\item We evaluate diverse vision-language agents and validate the evaluation protocol through agricultural expert agreement and sampled trajectory inspection, showing that \textsc{AgriGym} supports fine-grained analysis of evidence-grounded agricultural consultation.
\end{itemize}

\begin{table*}[t]
\centering
\scriptsize
\setlength{\tabcolsep}{3.2pt}
\renewcommand{\arraystretch}{1.15}

\resizebox{\textwidth}{!}{
\begin{tabular}{llcccccc}
\toprule
\textbf{Work} &
\textbf{Primary Focus} &
\makecell{\textbf{Real}\\\textbf{Consultation}\\\textbf{Cases}} &
\makecell{\textbf{External}\\\textbf{Evidence}\\\textbf{Layer}} &
\makecell{\textbf{Agentic}\\\textbf{Execution}} &
\makecell{\textbf{Tool}\\\textbf{Use}} &
\makecell{\textbf{Interactive}\\\textbf{Environment}} &
\makecell{\textbf{Trajectory-}\\\textbf{Level}\\\textbf{Evaluation}} \\
\midrule

CROP~\citep{crop} 
& Instruction tuning
& \xmark & \xmark & \xmark & \xmark & \xmark & \xmark \\

AgriBench~\cite{agribench} 
& Multimodal understanding 
& \xmark & \xmark & \xmark & \xmark & \xmark & \xmark \\

AgMMU~\cite{agmmu} 
& Knowledge-intensive VQA 
& \pmark & \xmark & \xmark & \xmark & \xmark & \xmark \\

AgroBench~\cite{agrobench} 
& Expert-annotated VLM evaluation 
& \xmark & \xmark & \xmark & \xmark & \xmark & \xmark \\

Agri-CM3~\cite{agri-cm3} 
& Multi-level reasoning 
& \xmark & \xmark & \xmark & \xmark & \xmark & \xmark \\

MIRAGE~\cite{mirage} 
& Expert-guided consultation 
& \cmark & \xmark & \xmark & \xmark & \pmark & \xmark \\

AgriGPT-VL~\cite{agrigpt-vl} 
& VLM and evaluation suite
& \pmark & \pmark & \pmark & \xmark & \xmark & \xmark \\

AgriEval~\cite{agrieval}
& \makecell[l]{Knowledge understanding\\ and open-ended answering}
& \xmark & \pmark & \xmark & \xmark & \xmark & \xmark \\

\midrule
\textbf{AgriGym} 
& \makecell[l]{\textbf{Executable evidence-grounded}\\ \textbf{agentic environment}}
& \cmark & \cmark & \cmark & \cmark & \cmark & \cmark \\

\bottomrule

\end{tabular}
}
\caption{Comparison between AgriGym and representative agricultural AI benchmarks. 
\cmark indicates that the feature is explicitly supported, \xmark indicates that it is not the main focus, and \pmark indicates partial or indirect support.}
\label{tab:benchmark_comparison}
\end{table*}

\section{Related Work}
\label{sec:related_work}
Prior work related to AgriGym falls into two lines.
\textbf{(1) Agricultural VLMs and benchmarks.}
Existing agricultural VLM systems ~\cite{agrigpt-rag,agrigpt-vl,agri-llava,agrogpt} and benchmarks~\cite{agrieval,agribench,agmmu,agrobench,agri-cm3,mirage,crop,cddm} evaluate multimodal understanding, crop and disease recognition, agricultural VQA, management recommendation, and consultation response generation.
MIRAGE is most related because it provides realistic consultation cases with user queries, images, metadata, and expert responses~\cite{mirage}.
However, these works primarily assess recognition, domain knowledge, or final response quality.
\textbf{(2) Tool-using and interactive agents.}
Recent agent benchmarks evaluate LLMs through tool use and executable environments~\cite{react,webvoyager,agentbench,webarena,toolbench,appworld,trajectbench}, but mainly focus on transactional or digital tasks such as API calling, web navigation, and application operation.
In contrast, AgriGym evaluates agricultural consultation as evidence-seeking tool use: agents must inspect symptoms, retrieve related cases, compare possible causes, and ground recommendations in a recorded evidence-seeking trajectory. Unlike agent benchmarks centered on task completion or tool execution, AgriGym evaluates whether evidence collected through a non-unique trajectory supports an open-ended agricultural recommendation. Related medical benchmarks also study interactive consultation,
information gathering, and tool use
~\cite{3mdbench,aihospital,matrix,meddialogrubrics}. A more detailed discussion and comparison are provided in Appendix~\ref{app:related}.

\begin{figure*}[t]
  \centering
  \includegraphics[width=\textwidth]{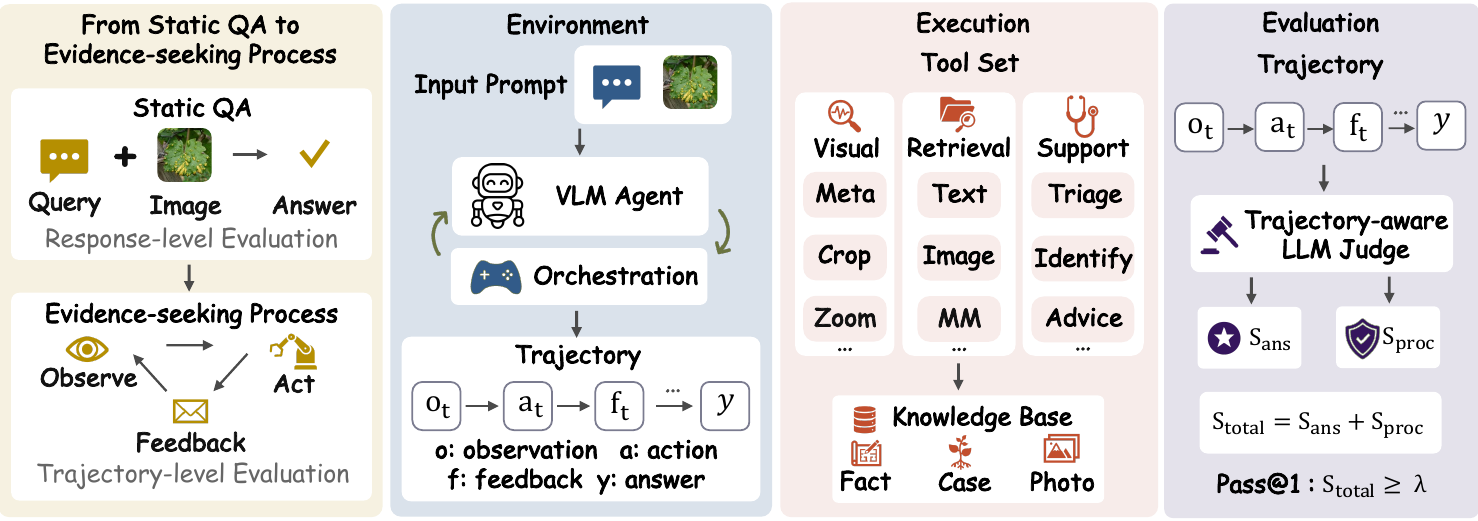}
  \caption{\textbf{Overview of AgriGym}. AgriGym converts static agricultural Q\&A into executable multi-step consultation episodes, where a VLM agent interacts with the environment, uses visual, retrieval, and domain-support tools, and produces a recorded evidence-seeking trajectory.
A trajectory-aware judge evaluates both answer quality and process reliability, allowing multiple valid tool-use paths rather than exact action matching.}
  \label{fig:model}
\end{figure*}

\section{Problem Reformulation: From Static Response Generation to Dynamic Evidence Seeking}
\label{sec:task}
Existing agricultural benchmarks typically treat an agricultural consultation case as a static Q\&A.
Given a user query $q$, a set of images $\mathcal{I}$, optional metadata $m$, and an expert reference answer $y^\star$, the model is asked to generate a final response $y$. The evaluation then focuses on whether $y$ is consistent with $y^\star$. This formulation does not expose the evidence-seeking process that leads to the recommendation. Therefore, AgriGym reformulates each consultation case as a finite evidence-seeking episode.
Let $\tau=(q,\mathcal{I},m,y^\star)$ denote a consultation task, where $y^\star$ is reserved only for evaluation. For each task, an agent $\pi_\theta$ interacts with an environment $\mathcal{E}_\tau$ initialized with the observable task context $(q,\mathcal{I},m)$. At step $t$, the agent receives an observation $o_t$ and selects an action according to its interaction history
\begin{equation}
h_t=(o_0,a_0,f_0,\ldots,o_t), \qquad
a_t \sim \pi_\theta(\cdot \mid h_t).
\end{equation}
In implementation, the full history $h_t$ can be represented by a compact trajectory summary for prompt construction, as described in Section~\ref{sec:agent_orchestration}. The action is either an evidence-seeking action or a final submission:
\begin{equation}
a_t \in \mathcal{A}_{\mathrm{env}} \cup \{\texttt{submit\_answer}\}.
\end{equation}
If $a_t\in\mathcal{A}_{\mathrm{env}}$, the environment returns feedback $f_t$ and the episode continues.
If $a_t=\texttt{submit\_answer}$, the agent outputs a recommendation $y$ and the episode terminates. The resulting trajectory is:
\begin{equation}
\Gamma=(o_0,a_0,f_0,\ldots,o_{T-1},a_{T-1},f_{T-1},a_T,y),
\end{equation}
where the final action $a_T=\texttt{submit\_answer}$ returns final answer $y$ and terminates the episode. A successful agent should not only produce a recommendation consistent with $y^\star$, but also construct a trajectory $\Gamma$ that provides sufficient support for $y$.
In this work, MIRAGE-MMST~\cite{mirage} provides the source consultation cases and expert reference answers. AgriGym preserves the original query, images, metadata, and reference answer, while adding the executable environment, available tools, interaction budget, and submission protocol. The resulting trajectories are generated by the evaluated agents rather than annotated by agricultural experts.

\section{AgriGym}
\subsection{Agent-Environment Interaction}
\label{sec:env}
To evaluate evidence seeking, AgriGym defines an executable environment that controls evidence access, returns structured feedback, and records action-feedback trajectories.
For each task $\tau$, the environment exposes $(q,\mathcal{I},m)$ to the agent while reserving $y^\star$ for evaluation. In the following section, we describe the two components that instantiate this execution layer: the tool set, which defines evidence-seeking actions, and the knowledge base, which supplies external evidence for retrieval tools.

\subsubsection{Tool Set}
\label{sec:tool_set}

The tool set defines the evidence-access actions available in the AgriGym environment.
The environment action space is decomposed into three groups:
\begin{equation}
\mathcal{A}_{\mathrm{env}}
=
\mathcal{A}_{\mathrm{vis}}
\cup
\mathcal{A}_{\mathrm{ret}}
\cup
\mathcal{A}_{\mathrm{sup}},
\end{equation}
where $\mathcal{A}_{\mathrm{vis}}$ contains visual inspection tools, $\mathcal{A}_{\mathrm{ret}}$ contains retrieval tools, and $\mathcal{A}_{\mathrm{sup}}$ contains domain-support tools. Specifically,
\begin{itemize}
    \item \textbf{Visual inspection tools} $\mathcal{A}_{\mathrm{vis}}$ operate on the input images and on visual outputs produced by previous visual tools, including metadata inspection, cropping, zooming, and bounding-box annotation. These tools allow the agent to focus on visual modality carefully.
\item \textbf{Retrieval tools} $\mathcal{A}_{\mathrm{ret}}$ access the external knowledge base (Section~\ref{sec:knowledge_base}).
They include \texttt{search\_knowledge\_base}, \texttt{visual\_retrieve}, \texttt{retrieve\_multimodal\_evidence}, and \texttt{retrieve\_by\_embedding}.
These tools return evidence that may support the consultation. 

\item \textbf{Domain-support tools} $\mathcal{A}_{\mathrm{sup}}$ return candidate causes and management references. They include \texttt{triage\_agri\_issue}, \texttt{identify\_disease}, \texttt{get\_management\_advice}.
These tools help route the issue type, retrieve candidate causes and management information.
\end{itemize}

All tool outputs are returned as observations, not oracle answers. The returned results are used by agents to determine the final recommendation. Each action records the agent's current state, the evidence gap being addressed, the rationale for the selected tool, and how returned evidence is accepted or rejected. These fields are stored in the trajectory for subsequent evaluation.

\subsubsection{Knowledge Base}
\label{sec:knowledge_base}

To support external evidence seeking, the retrieval tools in AgriGym are backed by a knowledge base constructed from existing agricultural facts and multimodal cases. We represent it as
\begin{equation}
\mathcal{K}=\{e_i\}_{i=1}^{N},
\qquad
e_i=(c_i,v_i,m_i),
\end{equation}
where $c_i$ is textual content, $v_i$ is an optional visual reference, and $m_i$ is metadata. Each entry is indexed in text and image embedding spaces:
\begin{equation}
z_i^{t}=\phi_t(c_i),
\qquad
z_i^{v}=\phi_v(v_i).
\end{equation}
During rollouts, the agent accesses $\mathcal{K}$ only through retrieval tools.

Given a text query $r_t$ and an image $u_t$, the retriever computes text, image, and multimodal similarities:
\begin{equation}
s_i^{t}=\cos\!\left(\phi_t(r_t),z_i^t\right), 
s_i^{v}=\cos\!\left(\phi_v(u_t),z_i^v\right),
\end{equation}
\begin{equation}
s_i^{mm}=\operatorname{Fuse}\!\left(s_i^t,s_i^v\right).
\end{equation}
For retrieval mode $m\in\{t,v,mm\}$, the returned evidence set is
\begin{equation}
\mathcal{R}_t^{m}
=
\operatorname{TopK}_{e_i\in\mathcal{K}} s_i^{m}.
\end{equation}

Retrieved items are evidence candidates rather than oracle answers: the agent must assess their relevance and decide whether to use them in the final recommendation. The knowledge base can be replaced or extended without changing the environment protocol. Construction details and the leakage audit are provided in
Appendices~\ref{app:kb} and~\ref{app:leakage}.

\subsection{Trajectory-Aware Evaluation}
\label{sec:judge}

Agricultural consultation \textbf{does not have a unique gold action sequence}. The expert reference answer defines the target outcome but does not specify a unique consultation trajectory under the AgriGym tool interface. AgriGym therefore evaluates the final recommendation together with the trajectory that produced it. We use an LLM-as-a-judge protocol~\cite{judge-arena,llm-as-examier,control}. 

The judge receives the task, expert reference answer, agent recommendation, executed actions, returned feedback, and cited evidence, and assigns two scores: $S_{\mathrm{ans}} \in[0,4]$ for semantic consistency with the expert reference and $S_{\mathrm{proc}} \in[0,6]$ for trajectory reliability. The protocol is judge-model agnostic, and our main experiments use GPT-4o. We further validate the judge against agricultural expert ratings in
Section~\ref{sec:human_validation}, examine judge-model sensitivity in
Table~\ref{tab:judge_sensitivity}, and analyze cross-judge consistency in
Section~\ref{sec:self_preference}.

The process score is defined as:
\begin{equation}
S_{\mathrm{proc}}
=
S_{\mathrm{ground}}
+
S_{\mathrm{traj}}
+
S_{\mathrm{tool}}
+
S_{\mathrm{vis}},
\end{equation}
where the four terms evaluate evidence grounding, trajectory quality, tool-use efficiency, and visual evidence use, respectively. The larger range of $S_{\mathrm{proc}}$ reflects its decomposition into four process dimensions, rather than an assumption that process quality should compensate for answer errors.

We additionally report the aggregate score
$S_{\mathrm{total}}=S_{\mathrm{ans}}+S_{\mathrm{proc}}$
for compact comparison, while interpreting answer and process scores separately.  The central goal of AgriGym is to evaluate answer quality and process reliability separately. This decomposition exposes answer--process mismatch: an agent may produce a broadly correct answer without a reliable supporting trajectory, or gather useful evidence but fail to synthesize a correct recommendation. The mismatch analysis in Section~\ref{sec:analysis} illustrates this motivation, while Section~\ref{sec:human_validation} evaluates the reliability of the trajectory-aware judge. The full scoring rubric and judge prompt are provided in Appendix~\ref{app:judge_prompt}.

\begin{table*}[!t]
\centering
\scriptsize
\setlength{\tabcolsep}{4.0pt}
\renewcommand{\arraystretch}{1.15}
\resizebox{\textwidth}{!}{
\begin{tabular}{lcccccccc}
\toprule
\textbf{Agent Model} &
$\mathbf{S_{\mathrm{total}}}$ &
$\mathbf{S_{\mathrm{ans}}}$ &
$\mathbf{S_{\mathrm{proc}}}$ &
$\mathbf{S_{\mathrm{ground}}}$ &
$\mathbf{S_{\mathrm{traj}}}$ &
$\mathbf{S_{\mathrm{tool}}}$ &
$\mathbf{S_{\mathrm{vis}}}$ &
Avg. Steps \\
\midrule

\multicolumn{9}{l}{\textit{Open-source VLM agents}} \\
\midrule
DeepSeek-VL2-Small        & 0.96 & 0.15 & 0.81 & 0.39 & 0.31 & 0.09 & 0.03 & 5.00 \\
Gemma-3-4B-IT             & 2.24 & 0.48 & 1.76 & 0.63 & 0.69 & 0.33 & 0.11 & 3.88 \\
Gemma-3-12B-IT           & 3.67 & 0.96 & 2.71 & 0.95 & 0.96 & 0.45 & 0.34 & 4.33 \\
InternVL3.5-2B-Instruct   & 3.27 & 1.17 & 2.10 & 0.74 & 0.74 & 0.30 & 0.32 & 4.25 \\
InternVL3.5-4B-Instruct   & 3.65 & 1.43 & 2.22 & 0.82 & 0.82 & 0.24 & 0.35 & 4.72 \\
Qwen3-VL-2B-Instruct      & 3.09 & 1.01 & 2.09 & 0.77 & 0.77 & 0.22 & 0.32 & 4.89 \\
Qwen3-VL-4B-Instruct      & 4.42 & 1.44 & 2.98 & 1.02 & 1.02 & 0.45 & 0.48 & 4.74 \\
Qwen3-VL-8B-Instruct      & 4.73 & 1.72 & 3.01 & 1.01 & 1.02 & 0.49 & 0.50 & 4.73 \\

\midrule
\multicolumn{9}{l}{\textit{Closed-source VLM agents}} \\
\midrule
GPT-4o                    & 5.18 & 1.70 & 3.48 & 1.12 & 1.12 & 0.64 & 0.60 & 3.66 \\
Gemini 3.1 Flash-Lite            & 7.03 & 2.63 & 4.41 & 1.44 & 1.43 & 0.90 & 0.63 & 2.45 \\
\bottomrule
\end{tabular}
}
\caption{Overall performance of multimodal agricultural agents on AgriGym. $S_{\mathrm{ans}}$ measures final recommendation quality, while $S_{\mathrm{proc}}$ measures the reliability of the supporting evidence trajectory. All the results are produced under the same settings.}
\label{tab:overall_results}
\end{table*}

\subsection{Agent Orchestration}
\label{sec:agent_orchestration}

The orchestrator is the runtime layer that converts VLM generations into valid AgriGym environment actions.
At step $t$, it builds the model input from the current observation $o_t$, a compact trajectory summary $M_t=\text{Summarize}(h_t)$, and the available action set:
\begin{equation}
x_t =
\text{Prompt}\!\left(
o_t,\,
M_t,\,
\mathcal{A}_{\mathrm{env}} \cup \{\texttt{submit\_answer}\}
\right).
\end{equation}
The generated output is parsed into a structured action, checked against schema and environment constraints, and then executed by the environment.
The orchestrator does not select tools, evidence, or recommendation content for the agent.
After each tool feedback, the summary is updated as
\begin{equation}
M_{t+1}=\text{Summarize}(h_{t+1}),
\end{equation}
so $M_t$ only compresses the recorded interaction history and does
not access the hidden expert answer or external task information.

To improve throughput, AgriGym executes multiple rollouts asynchronously across episodes.
Each episode remains internally sequential, but different episodes may independently wait for inference, execute tools, update summaries, or terminate.
This preserves the benchmark protocol while making multi-step agent evaluation scalable. We also provide a straightforward implementation of using GRPO~\cite{grpo} to train an agent on AgriGym with details in Appendix~\ref{sec:pilot_training} and Table~\ref{tab:pilot_training}. This pilot demonstrates the compatibility with agent training.

\section{Experiments}
\label{sec:experiments}
\paragraph{Model. }We evaluate ten open- and closed-source VLM agents, including DeepSeek-VL2-Small~\cite{ds-vl2}, Gemma-3~\cite{gemma3}, InternVL3.5~\cite{internvl3.5}, Qwen3-VL~\cite{Qwen3-VL}, GPT-4o~\cite{gpt4o}, and Gemini 3.1 Flash-Lite~\cite{gemini31_flash_lite}.

\paragraph{Dataset. }AgriGym is evaluated on MIRAGE-MMST~\cite{mirage}. Each case contains a user query, images, metadata, and an expert reference answer; the observable information is provided to the agent, while the expert answer is hidden and used only for evaluation. 

\paragraph{Metrics. }We primarily report answer quality $S_{\mathrm{ans}}$ and process reliability $S_{\mathrm{proc}}$ separately, together with $S_{\mathrm{total}}$, its process subscores, and average steps. Full implementation details are provided in Appendix~\ref{app:implementation}.

All agents use the same environment protocol, tool set, knowledge base, maximum step budget, and scoring rubric.

\subsection{Overall Results}
\label{sec:overall}
Table~\ref{tab:overall_results} shows that current VLM agents remain limited in both final-answer quality and process reliability. Most open-source agents obtain low continuous $S_{\mathrm{ans}}$ and $S_{\mathrm{proc}}$ scores, indicating substantial limitations in producing accurate recommendations supported by reliable evidence-seeking trajectories.
Among open-source models, Qwen3-VL-8B-Instruct performs best, and scaling within the Qwen3-VL and Gemma-3 families generally improves both $S_{\mathrm{ans}}$ and $S_{\mathrm{proc}}$. Closed-source agents achieve the strongest results.
Gemini 3.1 Flash-Lite obtains the best overall performance across $S_{\mathrm{total}}$, $S_{\mathrm{ans}}$, $S_{\mathrm{proc}}$, while requiring the fewest average steps.
Together with its high tool-efficiency score, this shows that strong performance does not require exhausting the interaction budget.

The results illustrate the diagnostic value of trajectory-aware evaluation. Answer scores alone obscure important differences in evidence grounding, trajectory quality, and tool-use behavior.
AgriGym exposes these differences by evaluating both final recommendation quality and the reliability of the supporting evidence-seeking process.

\begin{table}[!t]
\centering
\scriptsize
\setlength{\tabcolsep}{4.0pt}
\renewcommand{\arraystretch}{1.15}
\resizebox{\linewidth}{!}{
\begin{tabular}{lcccccccc}
\toprule
\textbf{Setting} &
$\mathbf{S_{\mathrm{total}}}$ &
$\mathbf{S_{\mathrm{ans}}}$ &
$\mathbf{S_{\mathrm{proc}}}$ &
$\mathbf{S_{\mathrm{ground}}}$ &
$\mathbf{S_{\mathrm{traj}}}$ &
$\mathbf{S_{\mathrm{tool}}}$ &
$\mathbf{S_{\mathrm{vis}}}$ &
\textbf{Avg. Steps} \\
\midrule
$\text{No tool use (response only)}$        & -- & 1.27 & -- & -- & -- & -- & -- & -- \\
$\text{Top-5 KB-Concat RAG}$        & -- & 1.31 & -- & -- & -- & -- & -- & -- \\
$\mathcal{A}_{\mathrm{vis}}$        & 4.10 & 1.39 & 2.71 & 0.96 & 0.95 & 0.14 & 0.66 & 4.97 \\
$\mathcal{A}_{\mathrm{ret}}$    & 4.31 & 1.57 & 2.74 & 0.96 & 0.95 & 0.36 & 0.48 & 4.17 \\
$\mathcal{A}_{\mathrm{sup}}$    & 3.41 & 1.19 & 2.22 & 0.93 & 0.83 & 0.21 & 0.25 & 4.47 \\
\midrule
AgriGym           & 4.42 & 1.44 & 2.98 &1.02 & 1.02 & 0.45 & 0.48  & 4.74 \\

\bottomrule
\end{tabular}
}
\caption{Static baselines and tool-group ablations on AgriGym using Qwen3-VL-4B-Instruct.}
\label{tab:tool_ablation}
\end{table}
\subsection{Ablation Study}
\label{sec:ablation}
We compare static baselines and tool-group ablations using Qwen3-VL-4B-Instruct under the same task split, knowledge base, and judge. Static RAG retrieves and concatenates the top-5 knowledge-base entries. The static baseline uses the same knowledge base and retriever as AgriGym, but performs retrieval only once before answer generation. As shown in Table~\ref{tab:tool_ablation}, static RAG yields only a small gain over no-tool generation, whereas retrieval-only achieves the highest answer score. The full AgriGym setting obtains the highest aggregate and process scores, indicating that interactive retrieval is more effective than fixed knowledge concatenation and that the complete tool set provides a more balanced process profile. We further evaluate the no-tool and static-RAG baselines with Qwen3-VL-2B and Qwen3-VL-8B to examine whether the comparison is specific to the 4B model. The additional results are reported in Appendix~\ref{app:static_baselines}.

To check judge variety, we also evaluate same Gemma-3-4B-IT trajectories with different judges. Table~\ref{tab:judge_sensitivity} shows that GPT-4o and Gemini 3.1 Flash-Lite give similar aggregate scores, while Qwen3-VL-8B assigns a lower process score; nevertheless, all judges assign low aggregate scores, indicating that the conclusion is not specific to a single judge. 

\begin{table}[!t]
\centering
\small
\setlength{\tabcolsep}{4.8pt}
\renewcommand{\arraystretch}{1.15}
\begin{tabular}{lccc}
\toprule
\textbf{Judge Model} &
$S_{\mathrm{ans}}$ &
$S_{\mathrm{proc}}$ &
$S_{\mathrm{total}}$ \\
\midrule
GPT-4o             & 0.48 & 1.76 & 2.24  \\
Gemini 3.1 Flash-Lite & 0.65 & 1.62 & 2.27  \\
Qwen3-VL-8B        & 0.67 & 1.28 & 1.96  \\
\bottomrule
\end{tabular}
\caption{
Sensitivity of LLM judge using Gemma-3-4B-IT as agent. 
}
\label{tab:judge_sensitivity}
\end{table}

\begin{figure}[!t]
    \centering
    \includegraphics[width=0.9\linewidth]{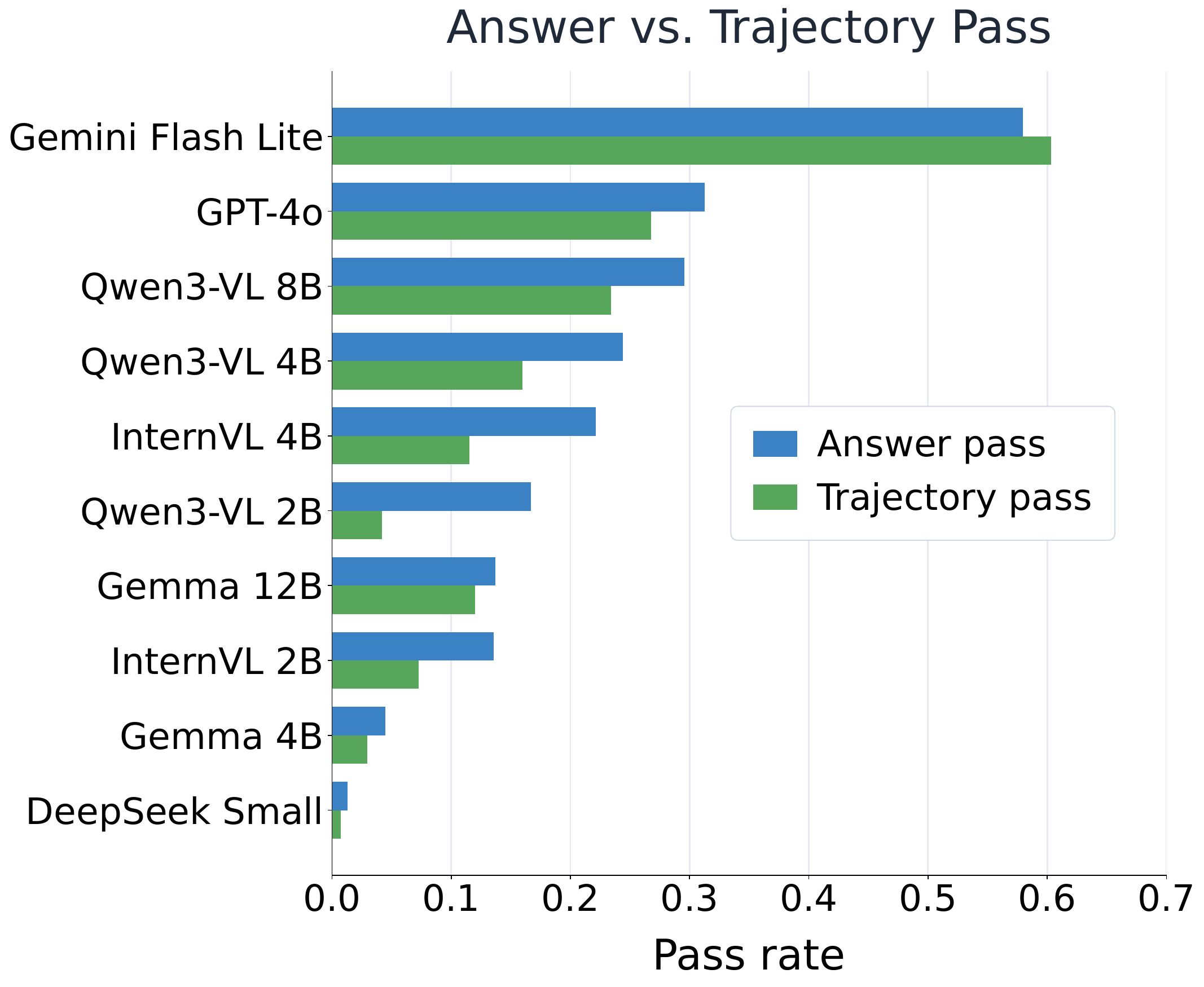}
    \caption{
    Answer quality and process reliability are separately scored aspects of agent performance.
    }
    \label{fig:answer_traj_pass}
\end{figure}

\begin{figure}[!t]
    \centering
    \includegraphics[width=0.9\linewidth]{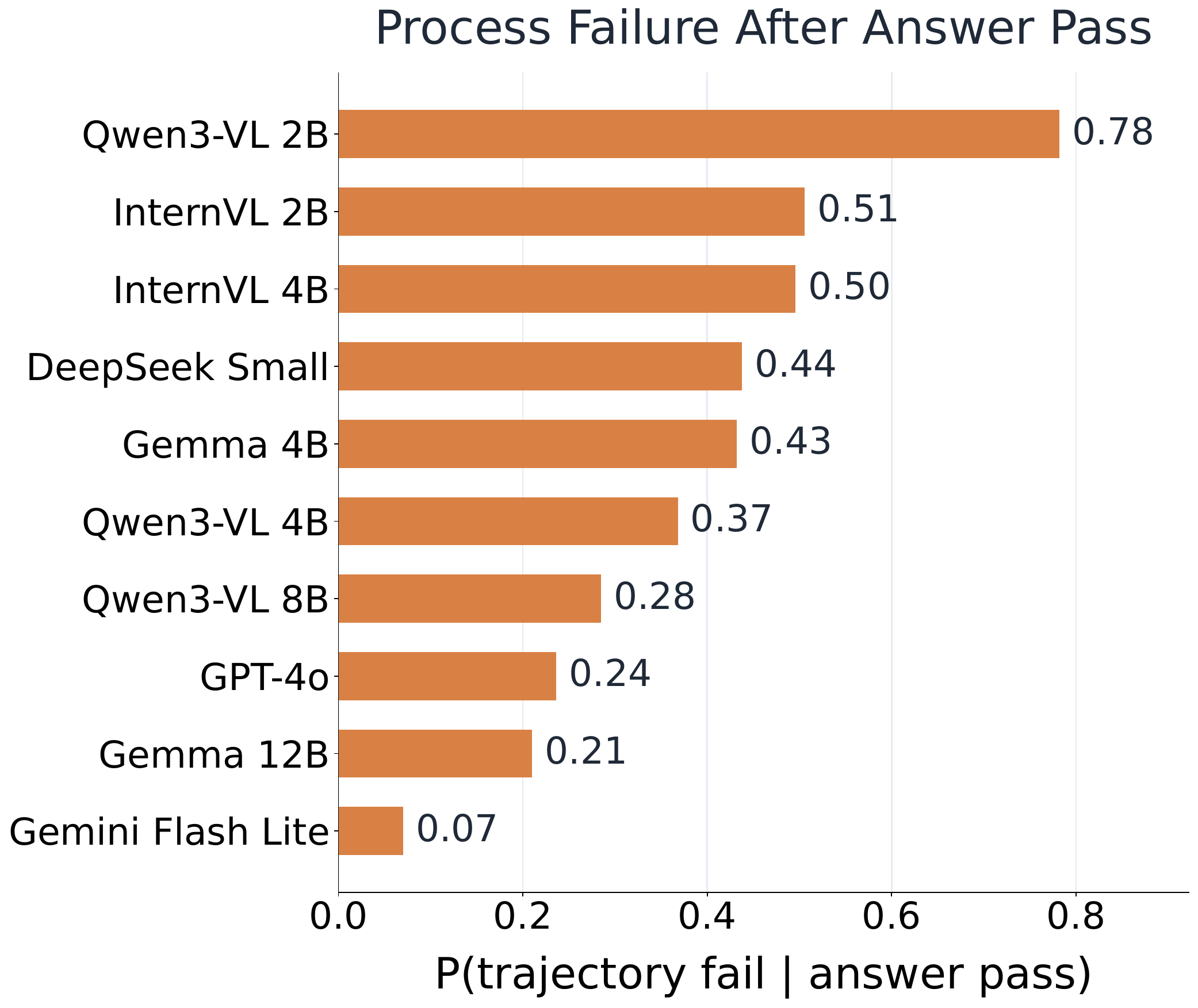}
    \caption{ \textbf{Answer unmatched with process}. $P(\text{TrajFail}|\text{AnsPass})$ measures how often an acceptable answer is not supported by a reliable trajectory. The results show that an acceptable answer is not always supported by a reliable process, validating the design of evaluation.
    }
    \label{fig:traj_fail_given_ans_pass}
\end{figure}

\subsection{Human Expert Validation}
\label{sec:human_validation}

We validate the trajectory-aware evaluation protocol on 40 trajectories
sampled across score distributions. Two agricultural experts independently
score the same trajectories using the evaluation rubric while remaining
blind to the LLM judge scores. We first evaluate agreement between the LLM
judge and an agricultural expert, and then examine inter-expert agreement
to assess the reliability of the rubric itself.

\paragraph{Expert--Judge Agreement.}
Table~\ref{tab:human_validation_main} reports the agreement between
Expert 1 and the LLM judge on the main evaluation dimensions.
The LLM judge shows strong rank correlation with expert assessment,
with Spearman correlations of 0.871, 0.903, and 0.815 for the total,
answer, and process scores, respectively.

\begin{table}[t!]
\centering
\small
\setlength{\tabcolsep}{4pt}
\begin{tabular}{lrrrr}
\toprule
Metric & Expert 1 & LLM & MAE & Spearman \\
\midrule
$\mathbf{S}_{\mathrm{total}}$  & 6.088 & 6.119 & 0.681 & 0.871 \\
$\mathbf{S}_{\mathrm{ans}}$    & 2.712 & 2.700 & 0.338 & 0.903 \\
$\mathbf{S}_{\mathrm{proc}}$   & 3.375 & 3.419 & 0.656 & 0.815 \\
$\mathbf{S}_{\mathrm{ground}}$ & 1.369 & 1.312 & 0.231 & 0.793 \\
$\mathbf{S}_{\mathrm{traj}}$   & 1.125 & 1.306 & 0.206 & 0.818 \\
\bottomrule
\end{tabular}
\caption{Agreement between Expert 1 and the LLM judge.}
\label{tab:human_validation_main}
\end{table}

\paragraph{Inter-Expert Agreement.}
To further assess the reliability of human evaluation, a second
agricultural expert independently scores the same 40 trajectories.
As shown in Table~\ref{tab:inter_expert_agreement}, the two experts
show strong agreement across all primary metrics, with Spearman
correlations of 0.91, 0.92, and 0.88 for total, answer, and process
scores, respectively. Across the process subscores, the inter-expert
Spearman correlations range from 0.75 to 0.89.

\begin{table}[t!]
\centering
\small
\setlength{\tabcolsep}{3.6pt}
\begin{tabular}{lrrrrr}
\toprule
Metric &
Expert 1 &
Expert 2 &
Pearson &
Spearman &
MAE \\
\midrule
$\mathbf{S}_{\mathrm{total}}$
& 6.09 & 5.80 & 0.93 & 0.91 & 0.70 \\

$\mathbf{S}_{\mathrm{ans}}$
& 2.71 & 2.65 & 0.91 & 0.92 & 0.26 \\

$\mathbf{S}_{\mathrm{proc}}$
& 3.38 & 3.15 & 0.87 & 0.88 & 0.56 \\
\bottomrule
\end{tabular}
\caption{Inter-expert agreement on the 40 human-evaluated trajectories.}
\label{tab:inter_expert_agreement}
\end{table}

Together, the expert-judge and inter-expert results indicate that the
trajectory-aware rubric produces consistent assessments across both
LLM-based and independent human evaluation.

\subsection{Cross-Judge Analysis}
\label{sec:self_preference}

To examine whether GPT-4o benefits from serving as both an evaluated agent and the default judge, we re-evaluate the trajectories generated by GPT-4o and Qwen3-VL-8B using Qwen3-VL-8B as an alternative judge, while keeping the agent outputs and scoring rubric unchanged. Based on Table~\ref{tab:self_preference}, the relative agent ordering remains unchanged under both judges.  We therefore do not observe evidence that GPT-4o systematically benefits from serving as both agent and judge.
\begin{table}[t]
\centering
\scriptsize
\setlength{\tabcolsep}{2.5pt}

\resizebox{\linewidth}{!}{
\begin{tabular}{llccc}
\toprule
\textbf{Agent} &
\textbf{Judge} &

$\mathbf{S_{\mathrm{total}}}$ &
$\mathbf{S_{\mathrm{ans}}}$ &
$\mathbf{S_{\mathrm{proc}}}$ \\
\midrule
GPT-4o
& GPT-4o
& 5.18 & 1.70 & 3.48 \\
GPT-4o
& Qwen3-VL-8B
& 5.35 & 1.85 & 3.50 \\
\midrule
Qwen3-VL-8B
& GPT-4o
& 4.73 & 1.72 & 3.01 \\
Qwen3-VL-8B
& Qwen3-VL-8B
& 4.82 & 2.02 & 2.80 \\
\bottomrule
\end{tabular}
}
\caption{Cross-judge evaluation of GPT-4o and Qwen3-VL-8B agents.}
\label{tab:self_preference}
\end{table}
We further conduct an expert inspection of 20 paired judge outputs for trajectories generated by Qwen3-VL-8B. The inspection indicates that Qwen3-VL-8B is generally more lenient on answer correctness but stricter on visual reasoning and tool-use efficiency. These differences are therefore more consistent with judge-specific calibration across rubric components than with GPT-4o self-preference.

\subsection{Analysis}
\label{sec:analysis}
\subsubsection{Answer-Process Relationship and Failure Analysis}
To quantify the additional information captured by trajectory-aware
evaluation, we analyze the relationship between final-answer quality and process reliability on all 8,188 Qwen3-VL-8B trajectories.
The two dimensions are related but not interchangeable:
$S_{\mathrm{ans}}$ and $S_{\mathrm{proc}}$ have a Spearman correlation
of 0.62 and a Pearson correlation of 0.64. For diagnostic analysis, we define answer adequacy as $S_{\mathrm{ans}}\geq3$ and process adequacy as $S_{\mathrm{proc}}\geq4$. Under these thresholds, 18.10\% of trajectories exhibit disagreement between answer and process assessments: 13.84\% obtain an adequate answer but an inadequate process, while 4.26\% show the opposite pattern. Thus, response-only evaluation cannot recover a substantial portion of the process-level information exposed by AgriGym. Figures~\ref{fig:answer_traj_pass} and
\ref{fig:traj_fail_given_ans_pass} further visualize these mismatches. Detailed failure patterns are provided in
Appendix~\ref{app:answer_process_failures}, with representative successful and near-miss trajectories in Appendices~\ref{app:successful_case} and~\ref{app:near_miss_case}.

\subsubsection{Tool Use Analysis}
\begin{figure}[t]
\centering
\includegraphics[width=\linewidth]{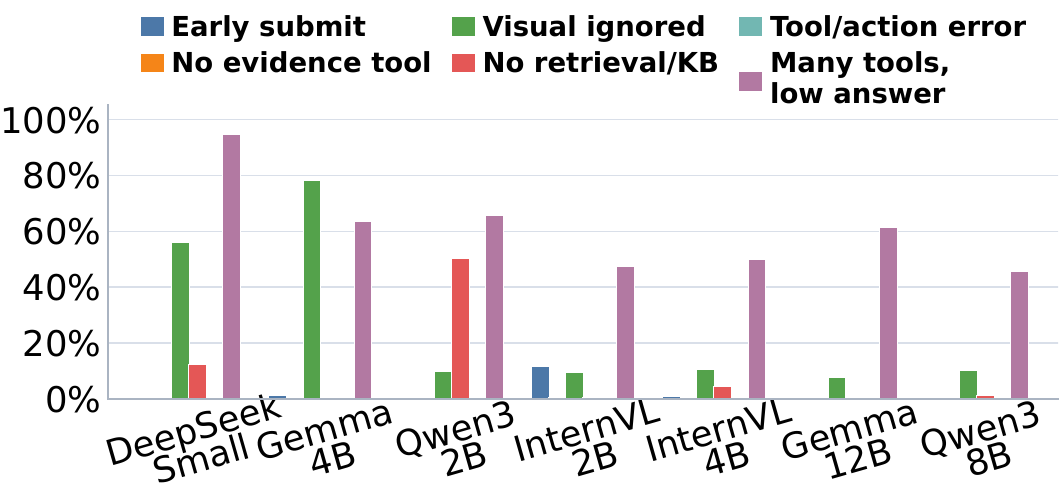}
\caption{
\textbf{Rule-based failure mode rates across agent models}. The results show that weaker agents fail for different reasons: some under-use evidence tools, while others call many tools without converting the evidence into accurate recommendations.
}
\vspace{-0.3cm}
\label{fig:failure_modes}
\end{figure}

\begin{figure}[t]
\centering
\includegraphics[width=\linewidth]{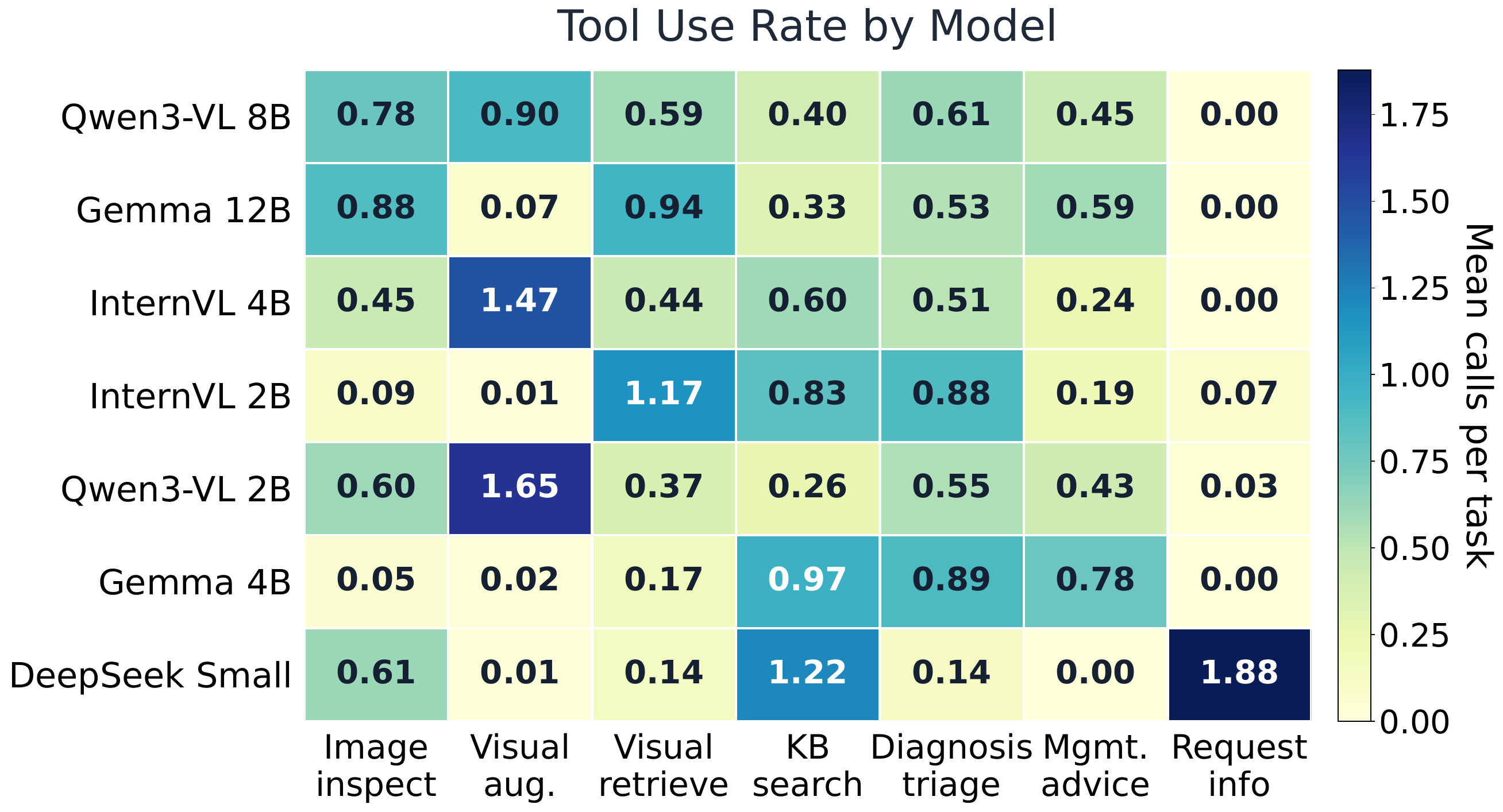}
\caption{\textbf{Tool-use profile across VLM agents}.
Stronger agents tend to distribute tool use across the tool set. This highlights that effective consultation depends on balanced evidence acquisition rather than the number of tool calls alone.}
\label{fig:tool_profile}
\end{figure}

Figures~\ref{fig:failure_modes} and~\ref{fig:tool_profile} diagnose tool-use behavior. Weak agents fail through different process patterns, including premature submission, missing retrieval evidence, and inefficient long trajectories. Stronger agents use a more balanced mixture of visual, retrieval, diagnostic, and management tools. This shows that reliable consultation depends on purposeful evidence acquisition rather than the number of tool calls alone.
Detailed definitions and statistics are provided in Appendix~\ref{app:tool_analysis}.

\subsection{Sensitivity Analysis}
\label{sec:sensitivity}

\paragraph{Step budget.} Table~\ref{tab:step_budget} compares $b\in\{3,5,7\}$ under the same task set, agent, tools, and judge.
Increasing the budget from $3$ to $5$ improves $S_{\mathrm{proc}}$ but $b=7$ gives no further gain. A budget of 7 increases interaction length without improving the aggregate score.

\begin{table}[!h]
\centering
\small
\setlength{\tabcolsep}{5.5pt}
\renewcommand{\arraystretch}{1.15}
\resizebox{\linewidth}{!}{

\begin{tabular}{ccccc}
\toprule
\textbf{Step Budget} &
$S_{\mathrm{ans}}$ &
$S_{\mathrm{proc}}$ &
$S_{\mathrm{total}}$ &
\textbf{Avg. Steps} \\
\midrule
$3$ & 1.68 & 2.63 & 4.31 & 2.98 \\
$5$ & 1.44 & 2.98 & 4.42 & 4.74 \\
$7$ & 1.55 & 2.86 & 4.41 & 6.04 \\
\bottomrule
\end{tabular}
}
\caption{Sensitivity to the maximum step budget $b$.}
\label{tab:step_budget}
\end{table}

\paragraph{Score aggregation.}
To examine whether the model ranking depends on the default score ranges, we normalize answer and process scores and vary their relative weighting:
\begin{equation}
S_{\lambda}
=
\lambda \frac{S_{\mathrm{ans}}}{4}
+
(1-\lambda)\frac{S_{\mathrm{proc}}}{6}.
\end{equation}
We test $\lambda\in\{0.25,0.50,0.75\}$.
The resulting rankings remain highly consistent with the default ranking, with Spearman correlations of 1.000, 0.988, and 0.964, respectively. The top four models are unchanged across all settings, indicating that the main conclusions are robust to alternative answer--process weighting.

\subsection{Trajectory Diversity and Qualitative Analysis}

We further examine whether AgriGym supports diverse evidence-seeking
strategies rather than inducing a fixed tool-use pattern.
Across all 8,188 trajectories generated by Qwen3-VL-8B, we observe
286 unique action sequences.
Domain-support tools appear in 94.5\% of trajectories, while
55.3\% use visual operations, 52.2\% use visual or multimodal retrieval,
43.9\% use management-support tools, 39.3\% access the knowledge base,
and 37.2\% invoke diagnostic tools.
These results show substantial variation in how agents combine visual,
retrieval, diagnostic, and management evidence during consultation.

Qualitative inspection further reveals distinct trajectory patterns, including successful evidence-grounded consultation, uncertainty-aware evidence seeking, answer-process mismatches, and inefficient trajectories with repeated or weakly informative tool calls. Detailed trajectory statistics are provided in ~\ref{app:trajectory_diversity}.

\subsection{Case Study}
Appendix~\ref{app:successful_case} and~\ref{app:near_miss_case} provide representative successful and near-miss trajectories. Together, they illustrate how AgriGym distinguishes answer quality from the reliability of the supporting trajectory: the successful case receives high scores because the final recommendation is correct and well supported. The near-miss case is penalized despite a mostly correct diagnosis because the management advice is unsupported.
This shows that final-answer correctness does not imply trajectory reliability. Trajectory-level evaluation can therefore identify failures in evidence grounding and recommendation that are hidden by answer-only scoring.

\section{Conclusion} \label{sec:conclusion} We introduced AgriGym, an executable benchmark for evidence-grounded agricultural consultation agents. AgriGym evaluates agricultural consultation as multi-step agent-environment interaction, where agents gather evidence, use tools, and ground final recommendations in recorded trajectories. Its trajectory-aware evaluation separates answer quality from process reliability. Human-expert agreement, cross-judge evaluation, and repeated-rollout analysis provide preliminary evidence for expert alignment and evaluation stability. The benchmark results reveal substantial gaps between final-answer quality and the reliability of the supporting evidence-seeking process across current VLM agents.

\section*{Acknowledgments}
 This research has been supported by Australian Research Council Projects (CE200100025, DP230101196 and DE250100919).

\section*{Limitations}
AgriGym is evaluated on MIRAGE-MMST under a fixed knowledge base, tool set, and step budget, so the findings may not fully generalize to other agricultural domains or evidence sources. We also do not estimate an expert performance ceiling under the same environment configuration.

Judge validation uses two agricultural experts on 40 stratified trajectories. Broader human and robustness evaluation remains valuable.

Finally, AgriGym is an evaluation benchmark rather than a deployment-ready advisory system; real-world use requires regional validation, safety checks, and expert oversight.

\newpage
\bibliography{custom}

\clearpage
\appendix

\section{Appendix}
\label{sec:appendix}

\subsection{Detailed Case Study}
\subsubsection{Successful Case: Maple Bladder Gall Consultation}
\label{app:successful_case}

We first present a successful trajectory to illustrate the intended evidence-seeking behavior in AgriGym.
The user asks about red and white bumps on maple leaves and whether treatment is required.
The reference answer identifies the issue as maple bladder gall caused by eriophyid mites and notes that the condition is generally harmless and does not require treatment.
The agent reaches the same conclusion and receives a high score of $9.84/10$, with $S_{\mathrm{ans}}=4.0/4$ and $S_{\mathrm{proc}}=5.84/6$.

\begin{figure}[t!]
\centering
\begin{tcolorbox}[
    width=\columnwidth,
    colback=green!3,
    colframe=green!45!black,
    boxrule=0.7pt,
    arc=1.8mm,
    left=1.0mm,
    right=1.0mm,
    top=0.8mm,
    bottom=0.8mm,
    title=\textbf{Case Study: Maple Bladder Gall Consultation},
    fonttitle=\bfseries\small
]
\scriptsize
\renewcommand{\arraystretch}{1.12}
\setlength{\tabcolsep}{3pt}

\begin{tabularx}{\columnwidth}{p{0.20\columnwidth}X}
\textbf{User query} &
What are the red and white bumps on these maple leaves, and do they require treatment? \\

\textbf{Reference} &
Maple bladder gall caused by eriophyid mites; generally harmless and no treatment is needed. \\

\textbf{Agent answer} &
\emph{The red and white bumps are likely maple bladder gall, which is generally harmless and requires no treatment.} \\

\textbf{Score} &
\textbf{9.84/10} ($S_{\mathrm{ans}}=4.0/4$, $S_{\mathrm{proc}}=5.84/6$). \\
\end{tabularx}

\vspace{0.4em}
\textbf{Agent workflow.}
\begin{enumerate}\itemsep2pt \parskip0pt \parsep0pt
    \item \texttt{inspect\_image\_metadata}: verifies that the image is usable for follow-up visual analysis.
    \item \texttt{draw\_bounding\_box} + \texttt{zoom\_image}: localizes and magnifies the leaf regions containing the bumps.
    \item \texttt{identify\_disease}: retrieves maple-related diagnostic evidence indicating a benign gall condition.
    \item \texttt{submit\_answer}: cites supporting KB evidence and recommends no treatment.
\end{enumerate}

\vspace{0.3em}
\textbf{Takeaway.}
The agent does not answer directly; it first gathers visual and external evidence, then grounds the final recommendation in the recorded trajectory.
\end{tcolorbox}
\caption{
Successful maple bladder gall trajectory.
The agent localizes the visible symptoms, retrieves relevant diagnostic evidence, and submits a grounded no-treatment recommendation.
}
\label{fig:case_study_maple}
\end{figure}

Figure~\ref{fig:case_study_maple} shows that the agent follows the intended consultation pattern.
It first verifies that the image can be analyzed, then localizes the abnormal regions on the leaf and zooms into the visible bumps.
After collecting visual evidence, the agent uses a disease-identification tool to retrieve maple-specific diagnostic evidence.
The final response correctly identifies maple bladder gall and recommends no treatment, matching the expert reference.
This case receives a high score because both the answer and the supporting trajectory are reliable: the diagnosis is correct, the visual evidence is relevant, and the final recommendation is explicitly supported by the retrieved evidence.

\subsubsection{Near-Miss Failure Case: Eastern Filbert Blight}
\label{app:near_miss_case}

We next present a near-miss failure case to show why final-answer correctness alone is insufficient.
The task asks the agent to diagnose a decades-old grafted hazelnut tree with recent limb dieback.
The reference answer identifies Eastern Filbert Blight caused by \textit{Anisogramma anomala}, but also states that chemical treatments are generally not recommended for mature trees with extensive damage.
The agent correctly identifies Eastern Filbert Blight, but its final recommendation incorrectly suggests applying fungicides.
The case receives $S_{\mathrm{total}}=6.5/10$, with $S_{\mathrm{ans}}=3.0/4$ and $S_{\mathrm{proc}}=3.5/6$.

\begin{figure}[t!]
\centering
\begin{tcolorbox}[
    width=\columnwidth,
    colback=red!2,
    colframe=red!55!black,
    boxrule=0.7pt,
    arc=1.8mm,
    left=1.0mm,
    right=1.0mm,
    top=0.8mm,
    bottom=0.8mm,
    title=\textbf{Near-Miss Failure: Eastern Filbert Blight},
    fonttitle=\bfseries\small
]
\scriptsize
\renewcommand{\arraystretch}{1.12}
\setlength{\tabcolsep}{3pt}

\begin{tabularx}{\columnwidth}{p{0.20\columnwidth}X}
\textbf{User query} &
A grafted hazelnut tree is several decades old and has recently developed limb dieback. The user asks whether this is a hazelnut disease. \\

\textbf{Reference} &
Eastern Filbert Blight caused by \textit{Anisogramma anomala}; prune dead and dying limbs, but chemical treatments are generally not recommended for mature trees with extensive damage. \\

\textbf{Agent answer} &
\emph{The hazelnut tree shows signs of Eastern Filbert Blight, causing limb dieback. Management includes pruning infected branches and applying fungicides.} \\

\textbf{Score} &
\textbf{6.5/10} ($S_{\mathrm{ans}}=3.0/4$, $S_{\mathrm{proc}}=3.5/6$). \\
\end{tabularx}

\vspace{0.3em}
\textbf{Failure reason.}
The agent identifies the disease correctly, but recommends fungicides despite the reference advice that chemical treatment is generally not recommended for a mature tree with extensive damage.
\end{tcolorbox}
\caption{
Near-miss failure case.
The agent correctly identifies Eastern Filbert Blight but gives unsupported fungicide advice, illustrating why management-specific grounding must be evaluated separately from disease identification.
}
\label{fig:case_study_filbert}
\end{figure}

Table~\ref{tab:near_miss_trace} shows the step-level trajectory.
The agent first checks image metadata, which provides little diagnostic value.
It then zooms into the image and draws a bounding box around the damaged region.
Although these visual actions help focus attention on bark damage, they do not provide sufficient management-specific evidence.
The disease-identification tool correctly suggests Eastern Filbert Blight, but the agent submits a final answer without retrieving or grounding the management recommendation.
As a result, the final response includes fungicide advice that is not supported by the reference answer.

\begin{table*}[t]
\centering
\scriptsize
\setlength{\tabcolsep}{4.0pt}
\renewcommand{\arraystretch}{1.15}
\begin{tabular}{clp{0.22\textwidth}p{0.36\textwidth}c}
\toprule
\textbf{Step} &
\textbf{Action} &
\textbf{Agent Rationale} &
\textbf{Judge Comment} &
\textbf{Step Score} \\
\midrule
1 &
\texttt{inspect\_image\_metadata} &
Check image metadata to confirm dimensions and format for planning next steps. &
Metadata check was unnecessary for disease identification. &
0.25 \\
2 &
\texttt{zoom\_image} &
Zooming helps observe fine details of bark damage for disease identification. &
Zooming helped observe bark damage but did not decisively aid identification. &
0.50 \\
3 &
\texttt{draw\_bounding\_box} &
Marking damaged area for focused analysis and future retrieval. &
Bounding box was not crucial for disease identification. &
0.25 \\
4 &
\texttt{identify\_disease} &
Directly identifies disease from symptom and species. &
Correctly identified disease but lacked management-advice specificity. &
0.75 \\
5 &
\texttt{submit\_answer} &
Submit final diagnosis and management recommendation. &
The answer was mostly correct, but included unsupported fungicide advice. &
0.50 \\
\bottomrule
\end{tabular}
\caption{
Step-level trace for the Eastern Filbert Blight near-miss failure.
The agent performs several visual and diagnostic actions, but fails to obtain sufficient management-specific evidence before submitting the final answer.
}
\label{tab:near_miss_trace}
\end{table*}

This case demonstrates a failure mode that answer-only evaluation may obscure.
A model can identify the disease correctly and still provide an unsuitable management recommendation.
In agricultural consultation, this distinction matters because the final advice is actionable.
AgriGym exposes this issue by evaluating the supporting trajectory, not only the disease label or final response.
The case therefore illustrates why correct identification is necessary but not sufficient for reliable agricultural consultation.

\subsection{Answer--Process Failure Patterns}
\label{app:answer_process_failures}

Answer quality evaluates whether the final conclusion is broadly correct,
whereas process reliability evaluates whether that conclusion is reached
and supported through appropriate evidence collection and tool use.
Consequently, an agent may obtain an adequate answer score while exhibiting
an inadequate supporting process. We observe three representative patterns:

\begin{itemize}
    \item \textbf{Correct conclusion without sufficient evidence.}
    The agent identifies the likely disease or cause but omits relevant
    visual inspection or knowledge retrieval needed to support the
    conclusion.

    \item \textbf{Correct conclusion through an unreliable process.}
    The agent reaches an appropriate conclusion after using irrelevant,
    repetitive, or weakly informative tools that provide little direct
    support for the final answer.

    \item \textbf{Correct diagnosis with unsupported downstream advice.}
    The agent identifies the disease correctly but provides treatment or
    management recommendations that are not supported by the collected
    evidence.
\end{itemize}

The Eastern Filbert Blight case in
Figure~\ref{fig:case_study_filbert} illustrates the third pattern. The
agent correctly identifies the disease, resulting in a broadly adequate
answer. However, it recommends fungicide treatment, whereas the expert
reference states that chemical treatment is generally not recommended for
a mature tree with extensive damage. The management recommendation is
neither supported by the collected evidence nor consistent with the expert
reference. This case demonstrates why a correct core diagnosis does not
necessarily imply a reliable consultation process.

\subsection{Additional Static Baselines}
\label{app:static_baselines}

\begin{table}[t]
\centering
\small
\setlength{\tabcolsep}{6pt}
\begin{tabular}{lcc}
\toprule
\textbf{Model} & \textbf{No Tool} & \textbf{Static RAG} \\
\midrule
Qwen3-VL-2B & 0.87 & 1.55 \\
Qwen3-VL-4B & 1.27 & 1.31 \\
Qwen3-VL-8B & 0.93 & 1.61 \\
\bottomrule
\end{tabular}
\caption{Additional response-only static baselines. Values report
$S_{\mathrm{ans}}$; static settings do not produce executable trajectories
and therefore do not receive process scores.}
\label{tab:additional_static_baselines}
\end{table}

The results show that the effect of static retrieval varies across model scales. More importantly, these response-only settings provide no executable trajectory for process-level evaluation, whereas AgriGym
evaluates both final-answer quality and the supporting evidence-seeking process.

\subsection{Detailed Human--LLM Agreement Analysis}
\label{app:human_validation}

We provide the complete score-level and decision-level agreement results for the 40 trajectories used in Section~\ref{sec:human_validation}. The expert used the same evaluation rubric as the LLM judge and remained blind to the model identity and LLM-generated scores.

\paragraph{Score-level agreement.}
Table~\ref{tab:human_validation_full} reports the expert and LLM means, bias, mean absolute error (MAE), and Spearman correlation for all scoring dimensions. Bias is defined as the expert score minus the LLM score.

\begin{table}[t]
\centering
\scriptsize
\setlength{\tabcolsep}{3.2pt}
\begin{tabular}{lrrrrr}
\toprule
Metric & Expert & LLM & Bias & MAE & Spearman \\
\midrule
$\mathbf{S}_{\mathrm{total}}$  & 6.088 & 6.119 & -0.031 & 0.681 & 0.871 \\
$\mathbf{S}_{\mathrm{ans}}$    & 2.712 & 2.700 &  0.013 & 0.338 & 0.903 \\
$\mathbf{S}_{\mathrm{proc}}$   & 3.375 & 3.419 & -0.044 & 0.656 & 0.815 \\
$\mathbf{S}_{\mathrm{ground}}$ & 1.369 & 1.312 &  0.056 & 0.231 & 0.793 \\
$\mathbf{S}_{\mathrm{traj}}$   & 1.125 & 1.306 & -0.181 & 0.206 & 0.818 \\
$\mathbf{S}_{\mathrm{tool}}$   & 0.456 & 0.362 &  0.094 & 0.219 & 0.689 \\
$\mathbf{S}_{\mathrm{vis}}$    & 0.425 & 0.438 & -0.013 & 0.163 & 0.754 \\
\bottomrule
\end{tabular}
\caption{Complete score-level agreement between the agricultural expert and the LLM judge.}
\label{tab:human_validation_full}
\end{table}

\paragraph{Decision-level agreement.}
We additionally convert the continuous scores into the pass/fail decisions used in our analysis. Table~\ref{tab:human_validation_decision} shows 80\%--90\% agreement between the expert and LLM judge.

\begin{table}[t]
\centering
\scriptsize
\setlength{\tabcolsep}{3.5pt}
\begin{tabular}{lrrrr}
\toprule
Criterion & Expert & LLM & Agree. & kappa \\
\midrule
$\mathbf{S}_{\mathrm{total}}\!\geq\!7$ & 0.425 & 0.425 & 0.800 & 0.591 \\
$\mathbf{S}_{\mathrm{ans}}\!\geq\!3$   & 0.600 & 0.550 & 0.900 & 0.796 \\
$\mathbf{S}_{\mathrm{proc}}\!\geq\!4$  & 0.375 & 0.375 & 0.850 & 0.680 \\
\bottomrule
\end{tabular}
\caption{Decision-level agreement between the agricultural expert and the LLM judge.}
\label{tab:human_validation_decision}
\end{table}

The agreement is strong for the primary scores and moderate to strong across the process subscores. They also yield identical pass rates for the total-score and process-score decisions. These results show that the trajectory-aware judge is broadly consistent with agricultural expert assessment at both the score and decision levels.

\subsection{Expert Inspection of High-Scoring Trajectories}
\label{app:expert_trajectory_inspection}

Separately from the human--LLM agreement study, an agricultural expert inspected 20 sampled high-scoring trajectories to assess whether they represented reasonable consultation processes. The expert judged 17 trajectories fully reasonable and the remaining 3 generally reasonable with minor process-level limitations. None was considered fundamentally unreasonable. This inspection supports the alignment of high trajectory scores with expert expectations, without treating any single expert action sequence as a unique gold trajectory.

\subsection{Related Work}
\label{app:related}
\subsubsection{Vision-Language Models and Benchmarks in Agriculture}
Large vision-language models~\cite{qwen2.5vl,Qwen3-VL, gemma3, gpt4} have increasingly been explored for agricultural consultation~\cite{agrigpt-rag,agrieval,agri-llava}. Recent work has introduced several agricultural benchmarks and datasets for evaluating domain knowledge and multimodal understanding~\cite{agribench,agmmu,agrobench,agri-cm3,mirage}.
For example, AgriBench proposes a hierarchical benchmark for multimodal LLMs in agriculture and introduces MM-LUCAS~\cite{agribench}, a multimodal dataset with agricultural landscape images, segmentation masks, depth maps, and metadata.
AgMMU~\cite{agmmu} builds a large-scale agricultural multimodal understanding benchmark from authentic grower--expert interactions, covering symptom recognition, species and pest identification, and management instructions.
Moreover, AgroBench~\cite{agrobench} evaluates VLMs over a broad range of crop and disease categories, with expert annotations across multiple agricultural VQA tasks.
Agri-CM3~\cite{agri-cm3} further studies Chinese multimodal agricultural understanding with multi-level reasoning tasks and large-scale evaluation of open-source and proprietary VLMs.
MIRAGE~\cite{mirage} is especially relevant to our work, as it provides realistic expert-guided agricultural consultation tasks with natural user queries, images, metadata, and expert responses.
These benchmarks have substantially advanced response-level evaluation for agricultural VLMs. 

Beyond evaluation, agriculture-specific models and systems have also been proposed. Recent agriculture-specific agent systems further explore multimodal
reasoning and tool orchestration, including AgriAgent and
AgriDoctor~\cite{agriagent,agridoctor}, while AgroTools evaluates tool-augmented multimodal agents in executable agricultural workflows~\cite{agrotools}. These works study agricultural agent capabilities and tool execution, whereas AgriGym focuses on whether an open-ended consultation recommendation is supported by the evidence collected along a
non-unique interaction trajectory.

General agent benchmarks evaluate whether an agent completes a task, reaches a target environment state, or executes the required tool operations. Agricultural consultation presents a different evaluation problem: the output is an open-ended recommendation, multiple evidence-seeking trajectories may be valid, and successful tool execution does not guarantee that the collected evidence supports the final claims. AgriGym therefore changes the evaluation target from task completion or tool correctness to evidence-to-recommendation consistency across the recorded interaction trajectory.

\subsubsection{Tool-using and Interactive Agents}
Recent work has increasingly studied language models as agents that reason, act, and interact with external environments~\cite{real, clawgym, clawbench,claweval,  webgym_ms}.
ReAct~\cite{react} introduced a general paradigm in which language models interleave reasoning traces with task-specific actions, allowing actions to gather information from tools or environments and reasoning to guide subsequent decisions.
Subsequent agent frameworks further explored feedback-driven improvement and open-ended environment interaction, such as MedGym~\cite{medgym} and Voyager~\cite{voyager}.
These works motivated a shift from evaluating isolated model outputs to evaluating action trajectories. AppWorld~\cite{appworld} provides an executable world of applications with hundreds of APIs and programmatic state-based evaluation. More recently, GAIA2~\cite{gaia2} evaluates LLM agents in an asynchronous environment which supports customization. TRAJECT-Bench further studies trajectory-aware evaluation for agentic tool use~\cite{trajectbench}. AgriGym shares the motivation of evaluating beyond final outcomes, but focuses on open-ended agricultural consultation, where multiple evidence-seeking paths may be valid and evaluation centers on whether the collected evidence supports the final recommendation. These benchmarks show the importance of evaluating agents through actions and trajectories rather than isolated outputs. However, their tools are often used to complete transactional or digital tasks, such as updating databases, operating applications, or following service policies. Agricultural consultation requires a different form of tool use: tools are used to acquire evidence, inspect visual symptoms, retrieve relevant cases, and ground a recommendation.

Interactive consultation has also been studied extensively in medical AI. AI Hospital~\cite{aihospital} and 3MDBench~\cite{3mdbench} simulate multi-turn clinical interactions, while MATRIX~\cite{matrix} emphasizes safety-oriented conversational evaluation and MedDialogRubrics~\cite{meddialogrubrics} evaluates information gathering and diagnostic reasoning with fine-grained rubrics. These studies demonstrate
the broader importance of interaction-aware evaluation, but operate on medical tasks, environments, and evaluation targets rather than agricultural evidence-grounded consultation.

Unlike prior agent benchmarks that mainly evaluate whether an agent can complete tasks, AgriGym evaluates whether an agent can construct an evidence trajectory for an open-ended agricultural consultation. This design shifts the evaluation target from simple tool execution to evidence-grounded problem solving process.

\subsection{Tool Schema}
\label{app:tool_schema}

AgriGym exposes a fixed set of registered tools to all evaluated agents.
Each tool is called through a structured action with an explicit tool name and parameter dictionary.
The environment executes the selected tool and returns a structured observation.
Table~\ref{tab:tool_schema} summarizes the tool groups used in our benchmark.

\begin{table*}[t]
\centering
\small
\begin{tabular}{p{0.18\linewidth}p{0.25\linewidth}p{0.47\linewidth}}
\toprule
\textbf{Group} & \textbf{Tools} & \textbf{Purpose} \\
\midrule
Visual inspection and augmentation
&
\texttt{inspect\_image\_metadata},
\texttt{crop\_image},
\texttt{zoom\_image},
\texttt{draw\_bounding\_box}
&
Support visual evidence acquisition by inspecting image properties, focusing on regions of interest, and creating visual artifacts that can be used in subsequent reasoning steps. \\
\midrule
Evidence retrieval
&
\texttt{search\_knowledge\_base},
\texttt{visual\_retrieve},
\texttt{retrieve\_multimodal\_evidence},
\texttt{retrieve\_by\_embedding}
&
Retrieve relevant agronomic facts, semantically similar cases, or multimodal evidence from the knowledge base. These tools provide external evidence that the agent must assess and integrate. \\
\midrule
Domain support
&
\texttt{triage\_agri\_issue},
\texttt{identify\_disease},
\texttt{get\_management\_advice}
&
Provide structured agricultural support for issue routing, candidate cause identification, and management option lookup. These tools are used as decision support rather than ground-truth oracles. \\
\bottomrule
\end{tabular}
\caption{Registered tool groups in AgriGym. Tool outputs are treated as observations for the agent to interpret, not as oracle answers.}
\label{tab:tool_schema}
\end{table*}

Each tool can be viewed as a function
\begin{equation}
g_i: (p_i, s_t) \rightarrow f_t,
\end{equation}
where $p_i$ denotes the tool parameters, $s_t$ is the current environment state, and $f_t$ is the structured feedback returned to the agent.
The feedback is appended to the trajectory and may later be cited by the agent in its final recommendation.

All tools follow the same interaction pattern.
The agent selects a tool action and supplies structured parameters.
The environment executes the tool and appends the returned observation to the trajectory.
For final evaluation, these observations are available to the trajectory-aware judge, allowing the judge to assess whether the final recommendation is grounded in evidence produced during the episode.

\subsection{Knowledge Base Construction}
\label{app:kb}

The retrieval layer in AgriGym is built from existing agricultural multimodal resources.
We process the source data into evidence units that can be used by the retrieval tools during agent rollout.
Each evidence unit contains textual information, optional visual references, metadata, and identifiers that allow retrieved evidence to be cited in the final answer and inspected by the judge.

We build both text and image retrieval indices.
The text index supports keyword and semantic search over processed agricultural facts.
The image index supports retrieval of visually similar cases from input images or visual artifacts generated during the episode.
For multimodal retrieval, text and image signals are combined to retrieve evidence relevant to both the user query and the observed visual symptoms.

During benchmark execution, the processed evidence base and embedding indices are loaded once before rollout.
The shared retriever is mounted into the tool registry and reused by all retrieval tools.
This ensures that all evaluated agents have access to the same evidence source and avoids reloading the retrieval model or embedding matrices for each task.
The retrieval outputs are returned to the agent as observations rather than oracle labels.

\begin{table}[t]
\centering
\footnotesize
\setlength{\tabcolsep}{4pt}
\renewcommand{\arraystretch}{1.12}
\begin{tabular}{p{0.30\columnwidth} p{0.60\columnwidth}}
\toprule
\textbf{Component} & \textbf{Description} \\
\midrule
Fact entries & Processed agricultural evidence units \\
T. embeddings & Semantic index for text-based retrieval \\
I. embeddings & Visual index for image-based retrieval \\
Text ID map & Maps text embeddings to evidence entries \\
Image ID map & Maps image embeddings to evidence entries \\
Metadata & Embedding model and index metadata \\
\bottomrule
\end{tabular}
\caption{Components of the knowledge base}
\label{tab:kb_components}
\end{table}

\subsection{Agent Training}
\label{sec:pilot_training}
Inspired by the recent research on agent with RL training~\cite{deepseek-r1,deepseekv3, agent-evolve, dapo, drgpro}, We conduct a pilot GRPO~\cite{grpo} experiment to examine whether the trajectory-aware scores designed in AgriGym can be used as reward signals for agent tuning.
We use Qwen3-VL-4B-Instruct as the base agent and construct the reward from all judge subscores, including final-answer quality, evidence grounding, trajectory quality, tool efficiency, and visual evidence use.
Since the reward and the evaluation metric are derived from the same framework, this experiment is intended as a proof of concept rather than a primary benchmark comparison.

Table~\ref{tab:pilot_training} shows that GRPO training version improves the aggregate performance of the agent. $S_{\mathrm{total}}$ increases from $4.42$ to $4.61$. The improvement is mainly driven by final-answer quality and tool-use efficiency: $S_{\mathrm{ans}}$ increases from $1.44$ to $1.74$, while $S_{\mathrm{tool}}$ increases from $0.45$ to $0.59$.
This suggests that judge-derived rewards can provide a useful optimization signal for improving selected agent capabilities, especially answer generation and tool-use efficiency.

However, the process score does not improve.
$S_{\mathrm{proc}}$ decreases from $2.98$ to $2.87$, despite the gains in $S_{\mathrm{ans}}$ and $S_{\mathrm{tool}}$.
This indicates that better final recommendations and more efficient tool use do not necessarily imply more reliable evidence-seeking trajectories.
In particular, optimizing the current reward can improve outcome-oriented behavior while leaving evidence grounding, trajectory quality, and visual evidence use under-optimized.
This result illustrates the diagnostic value of AgriGym: its decomposed scores reveal which capabilities improve under reward tuning and which remain bottlenecks.
It also suggests an important future direction: designing training objectives that jointly optimize answer quality, tool efficiency, and faithful evidence acquisition, rather than relying on a scalar reward alone.

\begin{table}[t]
\centering
\scriptsize
\setlength{\tabcolsep}{3.8pt}
\renewcommand{\arraystretch}{1.15}
\resizebox{\linewidth}{!}{
\begin{tabular}{lccccc}
\toprule
\textbf{Agent} &

$S_{\mathrm{total}}$ &
$S_{\mathrm{ans}}$ &
$S_{\mathrm{proc}}$ &
$S_{\mathrm{tool}}$ &
\textbf{Avg. Steps} \\
\midrule
Qwen3-VL-4B & 4.42 & 1.44 & 2.98 & 0.45 & 4.74 \\
GRPO-tuned   & 4.61 & 1.74 & 2.87 & 0.59 & 4.97 \\
\bottomrule
\end{tabular}
}
\caption{
Pilot GRPO tuning results using judge-derived rewards.
GRPO improves aggregate performance, final-answer quality, and tool-use efficiency, but does not improve overall process reliability.
}
\label{tab:pilot_training}
\end{table}

\subsection{Implementation Details}
\label{app:implementation}
The use of code and datasets are released with documents of existing datasets and model artifacts according to their public access conditions.

\paragraph{Framework structure.}
AgriGym is organized into modular components for task loading, environment execution, agent rollout, tool execution, model backend management, and judging.
The task loader converts MIRAGE-MMST samples into internal task objects.
The environment provides the \texttt{reset} and \texttt{step} interface for executable episodes.
The agent module builds prompts, parses structured actions, maintains an evidence ledger, and repairs invalid actions when needed.
The rollout module connects the agent and environment, records the trajectory, and constructs judge-readable context.
The judge module performs trajectory-aware scoring and computes deterministic aggregate metrics.
This separation keeps task execution, model inference, and evaluation independent. 

\paragraph{Model backends.}
Both the agent and the judge can use different model backends.
For the agent, AgriGym supports OpenAI-compatible vLLM endpoints, remote API backends, and local Transformers models.
The judge can also use remote APIs, a separate vLLM endpoint, or a local Transformers model.

\paragraph{Experimental parameters.}
Unless otherwise specified, all benchmark experiments are run on the MIRAGE-MMST test split with \texttt{MMST\_Standard}.
We use a fixed sampling seed of $42$ and the seeded sampling mode for sampled experiments; full-test experiments use all $8{,}188$ test examples.
Each agent episode has a maximum budget of $5$ environment steps.
The step budget is an upper bound rather than a required number of tool calls: an agent may terminate earlier by calling \texttt{submit\_answer}.
We use a batch size of $32$, up to $16$ concurrent agent trajectories, and up to $16$ concurrent judge requests and the agent keeps at most $8$ history messages in compact context.

For local open-source model inference, we use vLLM with bfloat16 precision with maximum model length $24{,}576$. The agent generation temperature is $0$ and the default maximum generation length is $4{,}096$ tokens, unless overridden for a specific model.
The judge uses deterministic decoding with temperature $0.0$.

The default knowledge base is the processed AGMMU-based KB embedding.
The shared embedding retriever is preloaded once before rollout and reused by all retrieval tools.
The embedding model recorded in the KB manifest is Qwen/Qwen3-VL-Embedding-8B.
The tool registry is serialized by default to avoid concurrent mutation of shared retrieval state.

\paragraph{Evaluation cost.}
A full local-agent evaluation requires approximately 7-10 GPU-hours per model on a single NVIDIA RTX PRO 6000 GPU. For API-based agents, full-test evaluation costs approximately USD 200-350 per model, while GPT-4o trajectory judging costs approximately USD 90-120 per full evaluation. These costs may vary with model and API pricing.

\paragraph{Retrieval and tool initialization.}
The framework loads a processed knowledge base and optionally preloads a shared embedding retriever before running agent trajectories.
Retrieval tools reuse this shared retriever during rollout, avoiding repeated embedding-index initialization for each task.
The tool registry is also shared across tasks, ensuring that all evaluated agents access the same evidence sources and tool schemas. The default multimodal retrieval tool retrieves both text candidates and image candidates using the shared embedding index, expands the candidate pool to at least $\max(4K,20)$ candidates, merges candidates by FAQ/case identifier, and returns the top-$K$ fused cases.
Unless otherwise specified, we use $K=5$.
For each merged candidate, the fused retrieval score is computed as a weighted sum of text similarity, image similarity:
\[
\text{Fuse} = 0.6 s^t + 0.4 s^v 
\]
Retrieved candidates are sorted by $s^{mm}$ and exposed to the agent as structured evidence; the hidden MIRAGE-MMST expert answer is never provided to the agent or retrieval tool.

\paragraph{Default configuration.}
Unless otherwise specified, the benchmark runs on the MIRAGE-MMST test split with a fixed sample seed.
The default agent backend is vLLM with \texttt{Qwen/Qwen3-VL-8B-Instruct}.
The default judge backend is an API judge with \texttt{gpt-4o}.
The default maximum number of agent steps is $5$.
The default batch size is $32$, with up to $16$ concurrent task trajectories and $16$ concurrent judge requests.
The default embedding model is Qwen3-vl-8b. Local open-source inference is conducted on a single NVIDIA RTX PRO 6000
GPU, while closed-source agents and API judges are accessed through their respective API endpoints.

\paragraph{Outputs and reproducibility.}
Outputs are grouped by agent--judge model pair.
For each run, the framework saves compact task-level results, agent trajectories, judge results, core trajectories for manual inspection, generated visual artifacts, logs, and a workflow summary.
The summary file records the dataset split, selected sample indices, model names, scheduler settings, judge configuration, metrics, tool usage statistics, and output paths.

More specifically, the trajectory summary is constructed deterministically from the recorded
action-observation history using the same template for all agents. Invalid
action repair only corrects formatting or schema violations and does not
change the selected tool or semantic arguments. A repaired action consumes
the original environment step. For reproducible comparison, we fix the sample seed, sample mode, task limit, model names, and judge configuration across runs.

\subsection{Outcome-Process Relationship}
\label{app:outcome_process}

\begin{figure}[t]
\centering
\includegraphics[width=\linewidth]{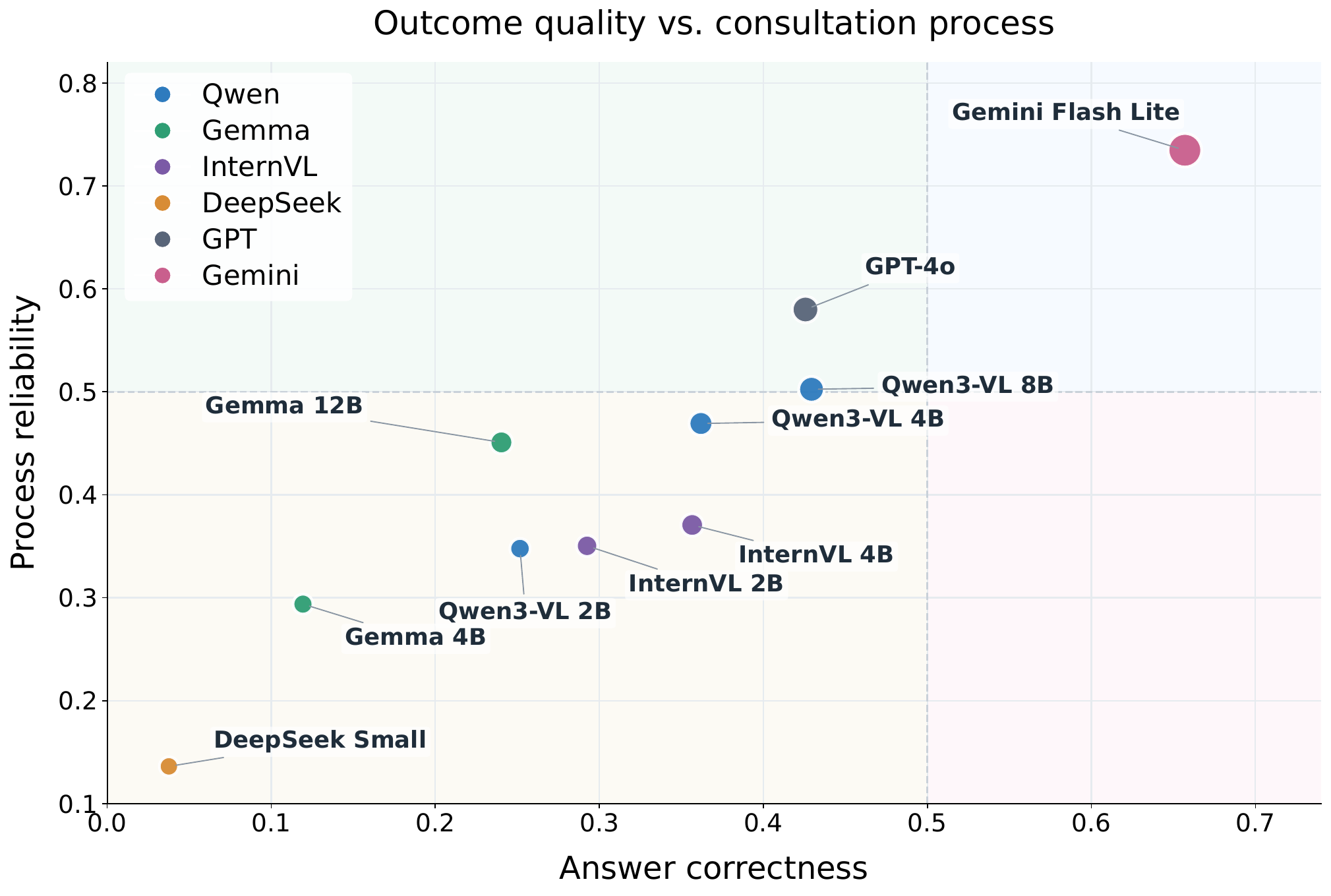}
\caption{
Outcome quality versus consultation-process reliability.
The x-axis shows normalized answer correctness, and the y-axis shows normalized process reliability.
Dashed lines mark midpoint thresholds for the two dimensions.
Models in the upper-right region produce both stronger final answers and more reliable supporting trajectories, while models below the horizontal line show weaker evidence-seeking processes.
The plot illustrates that answer quality and process reliability are related but not interchangeable, motivating the decomposed evaluation in AgriGym.
}
\label{fig:outcome_process}
\end{figure}

Figure~\ref{fig:outcome_process} visualizes the relationship between final-answer quality and consultation-process reliability across evaluated agents.
The x-axis reports normalized answer correctness, while the y-axis reports normalized process reliability.
The dashed lines mark the midpoints of the two axes and divide agents into four qualitative regions.

The upper-right region contains agents that achieve both strong final answers and reliable supporting trajectories.
Gemini 3.1 Flash-Lite is the only model clearly located in this region, indicating that its recommendations are more consistently supported by the consultation process.
GPT-4o also shows strong process reliability, but its answer correctness remains lower than Gemini 3.1 Flash-Lite.
Qwen3-VL-8B and Qwen3-VL-4B occupy the middle region, showing moderate answer quality with only partial process reliability.
This suggests that scaling within the Qwen3-VL family improves both dimensions, but still does not close the gap to stronger closed-source agents.

The lower-left and lower-middle regions contain most open-source agents.
Gemma-4B, InternVL-2B/4B, Qwen3-VL-2B, and DeepSeek Small all show limited process reliability, even when answer correctness is non-trivial.
This pattern confirms that final-answer quality and process reliability are not interchangeable: a model may improve its final response without producing a sufficiently grounded evidence-seeking trajectory.
For example, InternVL-4B has higher answer correctness than Gemma-12B, but lower process reliability, while Gemma-12B shows the opposite trend.
Such differences would be obscured by answer-only evaluation.

Overall, this analysis supports the main design of AgriGym.
Agricultural consultation agents should not be ranked only by final-answer correctness, because the reliability of the supporting trajectory varies substantially across models.
By separating answer quality from process reliability, AgriGym reveals whether a recommendation is merely plausible or actually supported by a coherent evidence-acquisition process.

\subsection{Trajectory Diversity and Representative Behaviors}
\label{app:trajectory_diversity}

Table~\ref{tab:trajectory_diversity} summarizes the coverage of
representative evidence-seeking behaviors across Qwen3-VL-8B trajectories. The categories are not mutually exclusive, as a trajectory may combine multiple evidence-seeking behaviors.
\begin{table}[H]
\centering
\small
\setlength{\tabcolsep}{6pt}
\begin{tabular}{lc}
\toprule
\textbf{Trajectory behavior} & \textbf{Coverage} \\
\midrule
Domain-support tools              & 94.5\% \\
Visual operations                 & 55.3\% \\
Visual/multimodal retrieval       & 52.2\% \\
Management support                & 43.9\% \\
Knowledge-base retrieval          & 39.3\% \\
Diagnostic tools                  & 37.2\% \\
Visual retrieval                  & 24.9\% \\
Triage                            & 23.6\% \\
Crop-related tools                &  2.2\% \\
\bottomrule
\end{tabular}
\caption{Coverage of representative evidence-seeking behaviors across
8,188 Qwen3-VL-8B trajectories.}
\label{tab:trajectory_diversity}
\end{table}

\subsection{Multiple-Rollout Evaluation}
\label{app:multiple_rollouts}
Besides the main result in Table~\ref{tab:overall_results}, to evaluate trajectory-sampling variance, we independently generate three additional trajectories for each test case using Qwen3-VL-8B-Instruct under the stochastic sampling strategy (Table~\ref{tab:multiple_rollouts_main}, ~\ref{tab:multiple_rollouts_subscores}). The original result is included for comparison but excluded from the mean and standard deviation. 

\begin{table*}[t]
\centering
\small
\setlength{\tabcolsep}{8pt}
\begin{tabular}{lcccc}
\toprule
Rollout

& $S_{\mathrm{total}}$
& $S_{\mathrm{ans}}$
& $S_{\mathrm{proc}}$
& Avg. Steps \\
\midrule
Original
& 4.73 & 1.72 & 3.01 & 4.73 \\
Rollout 1
& 4.62 & 1.69 & 2.93 & 4.74 \\
Rollout 2
& 4.79 & 1.76 & 3.03 & 4.74 \\
Rollout 3
& 4.69 & 1.59 & 3.10 & 4.74 \\
\midrule
\textbf{Mean $\pm$ Std}
& $\mathbf{4.70 \pm 0.085}$
& $\mathbf{1.68 \pm 0.085}$
& $\mathbf{3.02 \pm 0.085}$
& $\mathbf{4.74 \pm 0.002}$ \\
\bottomrule
\end{tabular}
\caption{Multiple-rollout results on the primary evaluation metrics.}
\label{tab:multiple_rollouts_main}
\end{table*}

\begin{table*}[t]
\centering
\small
\setlength{\tabcolsep}{3.5pt}
\caption{Process subscores across multiple rollouts.}
\label{tab:multiple_rollouts_subscores}
\begin{tabular}{lcccc}
\toprule
Rollout
& $S_{\mathrm{ground}}$
& $S_{\mathrm{traj}}$
& $S_{\mathrm{tool}}$
& $S_{\mathrm{vis}}$ \\
\midrule
Original
& 1.01 & 1.02 & 0.49 & 0.50 \\
Rollout 1
& 1.00 & 1.01 & 0.47 & 0.45 \\
Rollout 2
& 1.02 & 1.03 & 0.51 & 0.47 \\
Rollout 3
& 1.05 & 1.07 & 0.51 & 0.47 \\
\midrule
\textbf{Mean $\pm$ Std}
& $\mathbf{1.02 \pm 0.025}$
& $\mathbf{1.04 \pm 0.031}$
& $\mathbf{0.50 \pm 0.023}$
& $\mathbf{0.46 \pm 0.012}$ \\
\bottomrule
\end{tabular}
\end{table*}

The three additional rollouts show low variance across both the primary metrics and process subscores. The originally reported result is also close to the multi-rollout mean, including $4.73$ versus $4.70$ for $S_{\mathrm{total}}$ and $3.01$ versus $3.02$ for $S_{\mathrm{proc}}$. This confirms that the original result is representative and that the model-level conclusions are robust to trajectory sampling. The aggregate scores under stochastic decoding remain close to the deterministic reference, indicating limited sensitivity to this decoding setting. 

\subsection{Tool-use Behavior Analysis}
\label{app:tool_analysis}

We provide a more detailed analysis of tool-use behavior in AgriGym.
The goal is to understand whether agent failures come from insufficient evidence acquisition, poor tool selection, or inefficient use of the available interaction budget.
We analyze two complementary views: rule-based failure modes and tool-call profiles.

\paragraph{Rule-based failure modes.}
Figure~\ref{fig:failure_modes} reports the frequency of several automatically identified failure patterns.
We define \textit{early submit} as cases where the agent terminates before collecting sufficient evidence.
\textit{No evidence tool} indicates that the trajectory contains no substantive evidence-acquisition action.
\textit{No retrieval/KB} indicates that the agent does not use retrieval or knowledge-base tools when external evidence is expected to support the consultation.
\textit{Tool/action error} captures invalid, malformed, or unexecutable actions.
Finally, \textit{many tools, low answer} captures trajectories with extensive tool use but low final-answer quality, indicating that the agent fails to convert collected evidence into a correct recommendation.

The results show that weak performance is not explained by a single failure type.
Some agents fail because they lack of evidence, submitting answers without using relevant visual or retrieval tools.
Others perform many tool calls but still obtain weak answer scores, suggesting that tool use alone is insufficient without effective evidence selection and synthesis.
This distinction is important: a model can fail either by not seeking evidence or by seeking evidence inefficiently and failing to integrate it into the final recommendation.

\paragraph{Tool-call profile.}
Figure~\ref{fig:tool_profile} shows the average number of calls per task for each tool category.
The profile reveals clear differences in how agents allocate their interaction budget.
Stronger agents tend to distribute actions across visual inspection, retrieval, diagnostic triage, and management-advice tools.
This balanced pattern indicates that they can combine image evidence with external factual or case evidence before producing a recommendation.
In contrast, weaker agents often concentrate on a narrow subset of actions.
Some overuse shallow actions, while others rely heavily on a small number of retrieval or diagnostic tools without sufficient visual grounding.

\paragraph{Implications.}
These analyses support two conclusions.
First, tool access by itself does not guarantee reliable consultation.
An agent must choose tools that address the current evidence gap and then use the returned feedback to support or revise its recommendation.
Second, different tools contribute different forms of evidence.
Visual tools support symptom localization and image grounding, retrieval tools provide external factual or case-based support, and domain-support tools help organize candidate causes and management options.
Reliable agricultural consultation therefore requires balanced and purposeful evidence acquisition rather than longer trajectories or more tool calls.
This finding further motivates the process-level metrics in AgriGym, which evaluate not only whether tools were called, but whether the resulting trajectory provides useful support for the final answer.

\subsection{Entity Overlap Audit}
\label{app:leakage}
\begin{table}[t]
\centering
\small
\setlength{\tabcolsep}{5pt}
\renewcommand{\arraystretch}{1.15}
\begin{tabular}{lcc}
\toprule
\textbf{Overlap Statistic} & \textbf{Strict} & \textbf{Relaxed} \\
\midrule
Matched samples / all samples & 35.15\% & 36.81\% \\
Matched samples / candidate samples & 37.30\% & 39.07\% \\
Matched plant-like samples & 49.35\% & 51.93\% \\
Matched unique candidates & 20.89\% & 22.17\% \\
\bottomrule
\end{tabular}
\caption{
Species/entity-name overlap between MIRAGE-MMST test metadata and the AGMMU-based knowledge base.
Strict matching normalizes case and punctuation; relaxed matching additionally removes parenthetical variants.
The overlap reflects shared agricultural entity names rather than shared expert consultation answers.
}
\label{tab:species_overlap}
\end{table}
We analyze whether the retrieval knowledge base could introduce data leakage into the AgriGym evaluation.
This concern is relevant because the evidence base is constructed from agricultural multimodal knowledge resources, while the evaluation episodes are derived from MIRAGE-MMST consultation cases.
A potential source of overlap is that both resources may mention the same plant, pest, or disease names.
We therefore quantify species/entity-name overlap between the MIRAGE-MMST test metadata and the AGMMU-based knowledge base.

\paragraph{Overlap definition.}
We extract metadata candidates from the MIRAGE-MMST test split using \texttt{entity\_name}, \texttt{entity\_scientific\_name}, and \texttt{entity\_common\_names}.
We compare these candidates with KB species names under two matching rules.
The \textit{strict} rule lowercases and normalizes punctuation while retaining parenthetical cultivar or variant text.
The \textit{relaxed} rule additionally removes parenthetical variants, so that, for example, \textit{apple tree} may match \textit{apple tree (honeycrisp)}.
We also report plant-like entities separately, because metadata entities may refer to plants, pests, or diseases.

\paragraph{Observed overlap.}
The MIRAGE-MMST test split contains $8{,}188$ samples, of which $7{,}715$ samples contain at least one metadata candidate.
Under strict matching, $2{,}878$ samples match a KB species name, corresponding to $35.15\%$ of all samples and $37.30\%$ of samples with metadata candidates.
Under relaxed matching, $3{,}014$ samples match a KB species name, corresponding to $36.81\%$ of all samples and $39.07\%$ of samples with metadata candidates.
For plant-like metadata candidates, strict matching covers $2{,}309/4{,}679$ samples ($49.35\%$), and relaxed matching covers $2{,}430/4{,}679$ samples ($51.93\%$).
At the unique-candidate level, only $2{,}450/11{,}730$ unique metadata candidates match KB species under strict matching ($20.89\%$), and $2{,}601/11{,}730$ match under relaxed matching ($22.17\%$).

\paragraph{Retrieved-answer overlap audit.}
We further audit the actual Top-5 KB-Concat RAG retrieval results on the full MIRAGE-MMST test split to measure whether retrieved KB passages directly expose the hidden expert answers.
For each test episode, we concatenate the retrieved KB fact fields and compare them with the hidden expert answer using lexical overlap after stopword removal.
This audit is intended as a text-leakage proxy rather than a human relevance annotation.
Retrieval succeeds for all $8{,}188$ test episodes, but no retrieved passage contains the full hidden expert answer as an exact substring.
Only $2/8{,}188$ cases ($0.02\%$) contain near-identical overlap under the longest-contiguous-overlap criterion.
The mean gold-answer token recall is $6.84\%$ for the top-1 retrieved entry and $14.83\%$ for the top-5 retrieved entries, while the mean retrieved-token precision proxy is $14.13\%$.
Only $45/8{,}188$ cases ($0.55\%$) achieve high gold-answer token recall.
These results suggest that the KB retrieval provides partial agricultural evidence, but it does not directly reveal the hidden MIRAGE-MMST expert answers.

\paragraph{Why species overlap is not answer leakage.}
We do not treat this species/entity overlap as data leakage for three reasons.
First, the overlap is at the level of public entity names, such as plant, pest, or disease names.
Such names are expected to recur across agricultural resources and do not by themselves reveal the expert recommendation for a consultation case.
Second, AGMMU and MIRAGE-MMST differ in task formulation.
AGMMU-style entries are knowledge-fact or identification-oriented examples, often asking questions such as what plant or object appears in an image.
In contrast, MIRAGE-MMST contains open-ended agricultural consultation cases answered by experts, where the required response may involve diagnosis, causal explanation, management advice, uncertainty, and context-specific recommendation.
Third, during AgriGym rollout, the agent does not receive the MIRAGE-MMST expert answer.
The knowledge base only provides retrieved evidence candidates through tools; the agent must decide whether the returned evidence is relevant and how it supports the final recommendation.

\paragraph{Implication for evaluation.}
This analysis suggests that the knowledge base may contain general background evidence about some entities appearing in the test cases, but it does not directly expose the expert-grounded consultation answers.
Indeed, such controlled access to external evidence is part of the intended evaluation setting: agricultural consultants are expected to compare observed cases with known plants, pests, diseases, and prior examples.
The key evaluation question in AgriGym is therefore not whether an entity name exists in the evidence base, but whether the agent can use retrieved evidence appropriately, avoid over-claiming from partial matches, and ground its final recommendation in the recorded trajectory. The observed entity overlap does not by itself constitute consultation-answer leakage, because shared species names are expected across agricultural resources. However, this entity-level audit does not rule out all forms of semantic or visual near-duplicate overlap.

\subsection{LLM-As-a-Judge Evaluation}

\subsubsection{Scoring Rubric}
\label{app:scoring_rubric}

The trajectory-aware judge evaluates each completed AgriGym episode using five subscores.
The first subscore measures final-answer quality, while the remaining four subscores measure the reliability of the supporting evidence-seeking process.
The total score is computed as
\begin{equation}
S_{\mathrm{total}}
=
S_{\mathrm{ans}}
+
S_{\mathrm{ground}}
+
S_{\mathrm{traj}}
+
S_{\mathrm{tool}}
+
S_{\mathrm{vis}},
\end{equation}
where
\begin{equation}
S_{\mathrm{proc}}
=
S_{\mathrm{ground}}
+
S_{\mathrm{traj}}
+
S_{\mathrm{tool}}
+
S_{\mathrm{vis}}.
\end{equation}
Thus, $S_{\mathrm{ans}}\in[0,4]$, $S_{\mathrm{proc}}\in[0,6]$, and $S_{\mathrm{total}}\in[0,10]$.
The larger range of $S_{\mathrm{proc}}$ reflects that process reliability is decomposed into multiple dimensions, not that process quality should compensate for an incorrect answer.

\begin{table*}[t]
\centering
\small
\setlength{\tabcolsep}{5pt}
\caption{Representative cases illustrating the distinction between evidence grounding and trajectory quality.}
\label{tab:metric_separation_cases}
\begin{tabularx}{\textwidth}{lccX}
\toprule
Task
& $S_{\mathrm{ground}}$
& $S_{\mathrm{traj}}$
& Analysis \\
\midrule
\#712615
& 2.0
& 1.0
& The agent correctly identifies acorn squash and provides harvest cues supported by the available evidence. However, it invokes disease- and management-oriented tools that are unnecessary for a simple identification and harvest query, reducing the purposefulness of the trajectory. \\

\#795447
& 0.0
& 1.0
& The agent follows a reasonable visual-inspection and retrieval sequence, but over-interprets weak evidence and recommends fungicide. The action sequence is reasonable, while the final management claim remains insufficiently supported. \\
\bottomrule
\end{tabularx}
\end{table*}
\paragraph{Answer correctness.}
$S_{\mathrm{ans}}\in[0,4]$ measures whether the final recommendation matches the expert reference.
The judge considers the main identification, causal explanation, and management recommendation.
A score of $4$ indicates that the response is correct and complete with respect to the reference.
A score around $3$ indicates that the main diagnosis or recommendation is mostly correct but contains a missing detail, minor overclaim, or partially unsupported advice.
A score around $2$ indicates partial correctness, such as identifying a plausible issue but missing an important cause or management decision.
A score around $1$ indicates that only a small part of the answer is relevant.
A score of $0$ indicates an incorrect or irrelevant final answer.

\paragraph{Evidence grounding.}
$S_{\mathrm{ground}}\in[0,2]$ measures whether the final claims are supported by the recorded evidence.
The judge checks whether the answer cites or uses observations, tool outputs, retrieved evidence, or domain-support feedback that actually support the diagnosis and recommendation.
A high score requires the final answer to be grounded in relevant evidence rather than unsupported assertion.
The judge penalizes overclaiming, using weak evidence as if it were conclusive, ignoring contradictory evidence, or making management recommendations that are not supported by the trajectory.

\paragraph{Trajectory quality.}
$S_{\mathrm{traj}}\in[0,2]$ measures whether the sequence of actions forms a coherent evidence-seeking process.
A high-quality trajectory uses tools purposefully, adapts to returned observations, and progressively reduces uncertainty before submitting the final answer.
The judge does not require a fixed gold action sequence, since multiple valid consultation paths may exist.
Instead, it evaluates whether the trajectory is reasonable given the available evidence.
Low scores are assigned to trajectories that are shallow, repetitive, poorly motivated, or disconnected from the final recommendation.

\paragraph{Tool-use efficiency.}
$S_{\mathrm{tool}}\in[0,1]$ measures whether the agent uses tools efficiently.
The judge rewards actions that contribute useful evidence and penalizes unnecessary, repeated, malformed, or irrelevant tool calls.
This score is not simply the inverse of trajectory length: a longer trajectory can still receive a reasonable score if the additional actions provide relevant evidence, while a short trajectory can receive a low score if it skips necessary evidence acquisition.

\paragraph{Visual evidence use.}
$S_{\mathrm{vis}}\in[0,1]$ measures whether the agent appropriately uses image evidence when the task contains visual information.
A high score requires the agent to inspect or reason about visual symptoms relevant to the user query.
The judge penalizes trajectories that ignore important image evidence, rely only on textual priors, or make visual claims unsupported by the image or visual-tool outputs.

\paragraph{Separate diagnosis and aggregate success.}
The judge scores answer quality and process reliability separately to diagnose different aspects of agent performance. This separation concerns their evaluation targets and does not imply that the two scores are statistically independent or cannot be combined for an aggregate assessment. A wrong final answer does not automatically imply an unreliable process if the agent uses reasonable tools, acknowledges uncertainty, and grounds its reasoning in the available evidence. Conversely, a broadly correct answer may still receive a low process score when it relies on weak evidence, irrelevant actions, or unsupported management advice. We therefore report $S_{\mathrm{ans}}$ and $S_{\mathrm{proc}}$ separately to identify such mismatches.

\paragraph{Separation and Relationship of Scoring Dimensions.}
The scoring dimensions are separately defined to capture different aspects of agent performance, but they are not assumed to be statistically independent or orthogonal. $S_{\mathrm{ans}}$ evaluates the semantic quality of the final recommendation, whereas $S_{\mathrm{proc}}$ evaluates the reliability of the process that produces it. Within $S_{\mathrm{proc}}$, $S_{\mathrm{ground}}$ measures whether the final claims are supported by the collected evidence, $S_{\mathrm{traj}}$ evaluates whether the action sequence is purposeful and adaptive, $S_{\mathrm{tool}}$ evaluates tool-use efficiency, and $S_{\mathrm{vis}}$ evaluates the effective use of visual evidence. These dimensions are naturally related, but capture distinguishable failure modes, as illustrated in Table~\ref{tab:metric_separation_cases}.

The opposite score patterns show that the rubric can distinguish insufficient support for final claims from weaknesses in the executed action sequence. The dimensions are therefore separately scored and diagnostically distinct, while remaining closely related components of the same consultation process.

\label{app:judge_prompt}
\begin{tcolorbox}[
    enhanced,
    breakable,
    colback=gray!3,
    colframe=black!65,
    boxrule=0.6pt,
    arc=1.2mm,
    left=1.2mm,
    right=1.2mm,
    top=1.0mm,
    bottom=1.0mm,
    title=\textbf{Trajectory-Aware Judge Prompt},
    fonttitle=\bfseries\small,
    coltitle=black,
    colbacktitle=gray!15,
    attach boxed title to top center={yshift=-1.5mm}, 
    boxed title style={
        boxrule=0.4pt,
        arc=1mm,
        colframe=black!60,
        colback=gray!12
    }
]
\scriptsize

\textbf{Instruction.}
You are a strict evaluator for an agricultural VLM-agent benchmark.
Judge whether tool evidence helped answer the user query, not whether the agent followed a fixed tool sequence.
Return compact JSON only.

\vspace{0.5em}
\textbf{Rubric.}
\begin{itemize}[leftmargin=1.4em,itemsep=1pt,topsep=2pt]
    \item \texttt{answer\_correctness}: Does \texttt{pred\_answer} match \texttt{gt\_answer} on the main identification, cause, and management decision?
    \item \texttt{evidence\_grounding}: Are final claims supported by tool evidence, visual observations, retrieved facts, or justified expert inference from similar cases?
    \item \texttt{trajectory\_quality}: Were tool calls purposeful, adaptive to observations, and reasonable given the available evidence?
    \item \texttt{tool\_efficiency}: Did the agent avoid unnecessary, repeated, or irrelevant tool calls?
    \item \texttt{visual\_reasoning}: For image tasks, did the agent inspect or use image evidence relevant to the query?
\end{itemize}

\vspace{0.5em}
\textbf{Important separation.}
\begin{itemize}[leftmargin=1.4em,itemsep=1pt,topsep=2pt]
    \item \texttt{answer\_correctness} is the outcome score. Do not use it to automatically lower \texttt{evidence\_grounding}, \texttt{trajectory\_quality}, \texttt{tool\_efficiency}, or \texttt{visual\_reasoning}.
    \item If retrieval lacks a directly matching case, a wrong final answer can still receive reasonable process subscores when the agent used tools sensibly, qualified weak evidence, and stated uncertainty.
    \item Penalize process only for process errors: irrelevant or repeated tools, overstated unsupported claims, ignored useful evidence, ignored images, or misuse of rejected evidence.
\end{itemize}

\vspace{0.5em}
\textbf{Required output.}
The judge returns compact JSON with the following fields:
\begin{itemize}[leftmargin=1.4em,itemsep=1pt,topsep=2pt]
    \item \texttt{answer\_correctness} $\in [0,4]$;
    \item \texttt{evidence\_grounding} $\in [0,2]$;
    \item \texttt{trajectory\_quality} $\in [0,2]$;
    \item \texttt{tool\_efficiency} $\in [0,1]$;
    \item \texttt{visual\_reasoning} $\in [0,1]$;
    \item a brief justification for each score.
\end{itemize}

\end{tcolorbox}

\subsection{Agent Action Prompt}
\label{app:agent_prompt_box}

\begin{tcolorbox}[
    enhanced,
    breakable,
    colback=gray!3,
    colframe=black!65,
    boxrule=0.6pt,
    arc=1.2mm,
    left=1.2mm,
    right=1.2mm,
    top=1.0mm,
    bottom=1.0mm,
    title=\textbf{Agent Action Prompt},
    fonttitle=\bfseries\small,
    coltitle=black,
    colbacktitle=gray!15,
    attach boxed title to top center={yshift=-1.5mm},
    boxed title style={
        boxrule=0.4pt,
        arc=1mm,
        colframe=black!60,
        colback=gray!12
    }
]
\scriptsize

\textbf{Instruction.}
You are an agricultural tool-using agent.
At each step, select exactly one allowed action and return JSON only.
The prompt provides the list of allowed actions, the tool descriptions, tool-use guidance, and action-specific schemas.

\vspace{0.5em}
\textbf{Required action schema.}
\begin{verbatim}
{
  "action": "<action_name>",
  "params": {...},
  "working_hypothesis": "current best diagnosis/cause,
  and confidence",
  "evidence_gap": "what is still missing or uncertain",
  "accepted_evidence": [
    "tool result id or observation: why accepted"
  ],
  "rejected_evidence": [
    "tool result id or observation: why rejected"
  ],
  "why_this_tool": "why this action is useful now",
  "expected_contribution":,
  "what this action should add to the trajectory"
}
\end{verbatim}

\vspace{0.5em}
\textbf{Required schema for \texttt{submit\_answer}.}
\begin{verbatim}
{
  "answer": str,
  "evidence_used": [
    {
      "source_type": "visual_tool|retrieval|management_tool|
                      knowledge_base|tool_feedback",
      "source_id": "tool name or returned KB entry id",
      "claim_supported": str
    }
  ],
  "confidence": "low|medium|high",
  "remaining_uncertainty": str
}
\end{verbatim}

\end{tcolorbox}
\end{document}